\documentclass[twoside,11pt]{article}

\usepackage{jmlr2e}
\usepackage{cite}
\usepackage{url}
\usepackage{graphicx}
\usepackage{subfigure}
\usepackage{indentfirst}
\usepackage{booktabs}

\ShortHeadings{Hyper-Parameter Optimization: A Review of Algorithms and Applications}{Yu and Zhu}
\firstpageno{1}

\begin{document}

\title{Hyper-Parameter Optimization: A Review of Algorithms and Applications}

\author{\name Tong\ Yu \email yutong01@inspur.com \\
       \addr Department of AI and HPC \\
       Inspur Electronic Information Industry Co., Ltd\\
       1036 Langchao Rd, Jinan, Shandong, China
       \AND
       \name Hong\ Zhu \email zhuhongbj@inspur.com \\
       \addr Department of AI and HPC \\
       Inspur (Beijing) Electronic Information Industry Co., Ltd\\
       2F, Block C, 2 Xinxi Rd., Shangdi, Haidian Dist, Beijing, China}

\editor{}

\maketitle

\begin{abstract}
Since deep neural networks were developed, they have made huge contributions to people’s everyday lives. Machine learning provides more rational advice than humans are capable of in almost every aspect of daily life. However, despite this achievement, the design and training of neural networks are still challenging and unpredictable procedures that have been alleged to be   “alchemy”. To lower the technical thresholds for common users, automated hyper-parameter optimization (HPO) has become a popular topic in both academic and industrial areas. This paper provides a review of the most essential topics on HPO. The first section introduces the key hyper-parameters related to model training and structure, and discusses their importance and methods to define the value range. Then, the research focuses on major optimization algorithms and their applicability, covering their efficiency and accuracy especially for deep learning networks. This study next reviews major services and tool-kits for HPO, comparing their support for state-of-the-art searching algorithms, feasibility with major deep-learning frameworks, and extensibility for new modules designed by users. The paper concludes with problems that exist when HPO is applied to deep learning, a comparison between optimization algorithms, and prominent approaches for model evaluation with limited computational resources.
\end{abstract}


\section{Introduction}
In the past several years, neural network techniques have become ubiquitous and influential in both research and commercial applications. In the past 10 years, neural networks have shown impressive results in image classification~\citep{szegedy16, he16}, objective detection~\citep{girshick15,redmon16}, natural language understanding~\citep{hochreiter97,vaswani2017attention}, and industrial control systems~\citep{abbeel16,hammond17}.However, neural networks are efficient in applications but inefficient when obtaining a model. It is thought to be a “brute force” method because a network is initialized with a random status and trained to an accurate model with an extremely large dataset. Moreover, researchers must dedicate their efforts to carefully coping with model design, algorithm design, and corresponding hyper-parameter selection, which means that the application of neural networks comes at a great price. “Based on experience” is generally the most widely used method, which means a practicable set of hyper-parameters requires researchers to have experience in training neural networks. However, the credibility of empirical values is weakened because of the lack of logical reasoning. In addition, experiences generally provide “workable” instead of “optimal” hyper-parameter sets.\\
\indent
The “no free lunch theorem” suggests that the computational cost for any optimization problem is the same for all problems and no solution offers a shortcut~\citep{wolpert97,igel14}.A feasible alternative for computational resources is the preliminary knowledge of experts, which is efficient in selecting influential parameters and narrowing down the search space. To save the rare resource of experts’ experience, automated machine learning (AutoML) has been proposed as a burgeoning technology to design and train neural networks automatically, at the cost of computational resources~\citep{feurer15,katz16,bello17,zoph16,jin18}.Hyper-parameter optimization (HPO) is an important component of AutoML in searching for optimum hyper-parameters for neural network structures and the model training process.\\
\indent
Hyper-parameter refers to parameters that cannot be updated during the training of machine learning. They can be involved in building the structure of the model, such as the number of hidden layers and the activation function, or in determining the efficiency and accuracy of model training, such as the learning rate (LR) of stochastic gradient descent (SGD), batch size, and optimizer~\citep{hyper}.The history of HPO dates back to the early 1990s~\citep{ripley93,king95},and the method is widely applied for neural networks with the increasing use of machine learning. HPO can be viewed as the final step of model design and the first step of training a neural network. Considering the influence of hyper-parameters on training accuracy and speed, they must carefully be configured with experience before the training process begins~\citep{rodriguez18}. The process of HPO automatically optimizes the hyper-parameters of a machine learning model to remove humans from the loop of a machine learning system. As a trade of human efforts, HPO demands a large amount of computational resources, especially when several hyper-parameters are optimized together. The questions of how to utilize computational resources and design an efficient search space have resulted in various studies on HPO on algorithms and toolkits. Conceptually, HPO’s purposes are threefold~\citep{feurer19}: to reduce the costly menial work of artificial intelligence (AI) experts and lower the threshold of research and development; to improve the accuracy and efficiency of neural network training~\citep{melis17}; and to make the choice of hyper-parameter set more convincing and the training results more reproducible~\citep{bergstra2013making}.\\
\indent
In recent years, HPO has become increasingly necessary because of two rising trends in the development of deep learning models. The first trend is the upscaling of neural networks for enhanced accuracy~\citep{tan19b}.Empirical studies have indicated that in most cases, more complex machine learning models with deeper and wider layers work better than do those with simple structures~\citep{he16,zagoruyko16,huang19}.The second trend is to design a tricky lightweight model to provide satisfying accuracy with fewer weights and parameters~\citep{ma18,sandler18,tan19a}. In this case, it is more difficult to adapt empirical values because of the stricter choices of hyper-parameters. Hyper-parameter tuning plays an essential role in both cases: a model with a complex structure indicates more hyper-parameters to tune, and a model with a carefully designed structure implies that every hyper-parameter must be tuned to a strict range to reproduce the accuracy. For a widely used model, tuning its hyper-parameters by hand is possible because “the ability to tune by hand” depends on experience and researchers can always borrow knowledge from previous works. This is similar for models at a small scale. However, for a larger model or newly published models, the wide range of hyper-parameter choices requires a great deal of menial work by researchers, as well as much time and computational resources for trial and error.\\

\indent
In addition to research, the industrial application of deep learning is a crucial practice in automobiles, manufacturing, and digital assistants. However, even for trained professional researchers, it is still no easy task to explore and implement a favorable model to solve specific problems. Users with less experience have substantial needs for suggested hyper-parameters and ready-to-use HPO tools. Motivated by both academic needs and practical application, automated hyper-parameter tuning services~\citep{golovin17,amazon18}and toolkits~\citep{liaw18,msr18}provide a solution to the limitation of manual deep learning designs.\\

\indent
This study is motivated by the prosperous demand for design and training deep learning network in industry and research. The difficulty in selecting proper parameters for different tasks makes it necessary to summarize existing algorithms and tools. The objective of this research is to conduct a survey on feasible algorithms for HPO, make a comparison on leading tools for HPO tasks, and propose challenges on HPO tasks on deep learning networks. Thus, this remainder of this paper is structured as follows. Section 2 begins with a discussion of key hyper-parameters for building and training neural networks, including their influence on models, potential search spaces, and empirical values or schedules based on previous experience. Section 3 focuses on widely used algorithms in hyper-parameter searching, and these approaches are categorized into searching algorithms and trial schedulers. This section also evaluates the efficiency and applicability of these algorithms for different machine learning models. Section 4 provides an overview of mainstream HPO toolkits and services, compares their pros and cons, and presents some practical and implementation details. Section 5 more comprehensively compares existing HPO methods and highlights the efficient methods of model evaluation, and finally, Section 6 provides the study’s conclusions.\\

\indent
The contribution of this study is summarized as follows:
\begin{itemize}
\item[-]Hyper-parameters are systematically categorized into structure-related and training-related. The discussion of their importance and empirical strategies are helpful to determine which hyper-parameters are involved in HPO.
\item[-]HPO algorithms are analyzed and compared in detail, according to their accuracy, efficiency and scope of application. The analysis on previous studies is not only committed to include state-of-the-art algorithms, but also to clarify limitations on certain scenarios.
\item[-]By comparing HPO toolkits, this study gives insights of the design of close-sourced libraries and open-sourced services, and clarifies the targeted users for each of them.
\item[-]The potential research direction regarding to existing problems are suggested on algorithms, applications and techniques.
\end{itemize}

\section{Major Hyper-Parameters and Search Space}
\indent
Considering the computational resources required, hyper-parameters with greater importance receive preferential treatment in the process of HPO. Hyper-parameters with a stronger effect on weights during training are more influential for neural network training~\citep{ng2017improving}. It is difficult to quantitatively determine which of the hyper-parameters are the most significant for final accuracy. In general, there are more studies on those with higher importance, and their importance has been decided by previous experience.\\
\indent
Hyper-parameters can be categorized into two groups: those used for training and those used for model design. A proper choice of hyper-parameters related to model training allows neural networks to learn faster and achieve enhanced performance. Currently, the most adopted optimizer for training a deep neural network is stochastic gradient descent~\citep{robbins51} with momentum~\citep{qian99} as well as its variants such as AdaGrad~\citep{duchi11}, RMSprop~\citep{hinton12b}, and Adam~\citep{kingma14}. In addition to the choice of optimizer, corresponding hyper-parameters (e.g., momentum) are critical for certain networks. During the training process, batch size and LR draw the most attention because they determine the speed of convergence, the tuning of which should always be ensured. Hyper-parameters for model design are more related to the structure of neural networks, the most typical example being the number of hidden layers and width of layers. These hyper-parameters usually measure the model’s learning capacity, which is determined by the complexity of function~\citep{agrawal18}.\\
\indent
This section provides an in-depth discussion on hyper-parameters of great importance for model structures and training, as well as introduce their effect in models and provide suggested values or schedules.

\subsection{Learning Rate}
\begin{figure}
    \centering
    \includegraphics{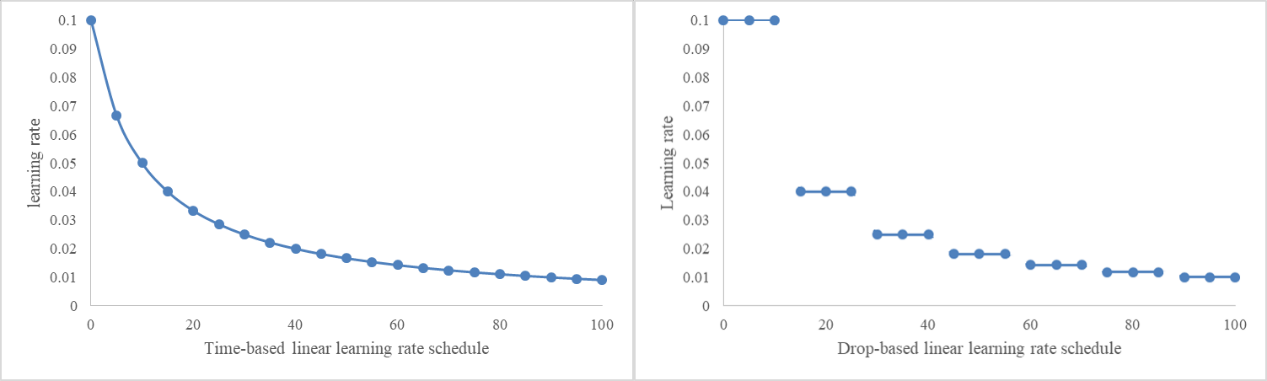}
    \caption{Linear decay of learning rate with time-based (left) and drop-based (right) schedules}
    \label{Figure 1}
\end{figure}

\begin{figure}
    \centering
    \includegraphics{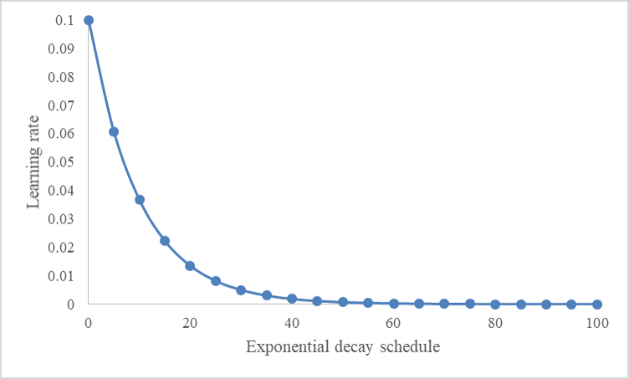}
    \caption{Exponential decay of learning rate}
    \label{Figure 2}
\end{figure}

\begin{figure}[htbp]
    \centering
    \includegraphics{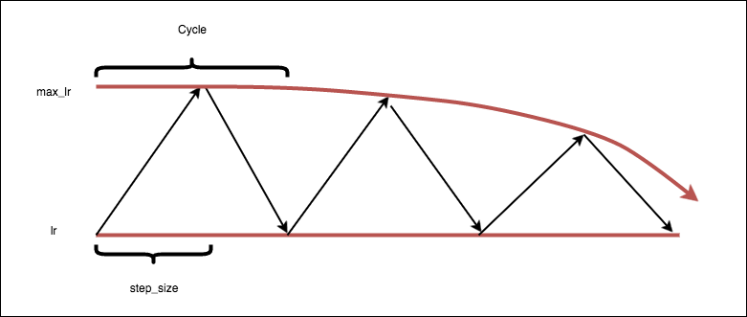}
    \caption{Learning rate decay in a cyclic schedule 
    (source: https://github.com/bckenstler/CLR)}
    \label{Figure 3}
\end{figure}

\begin{figure}
    \centering
    \includegraphics{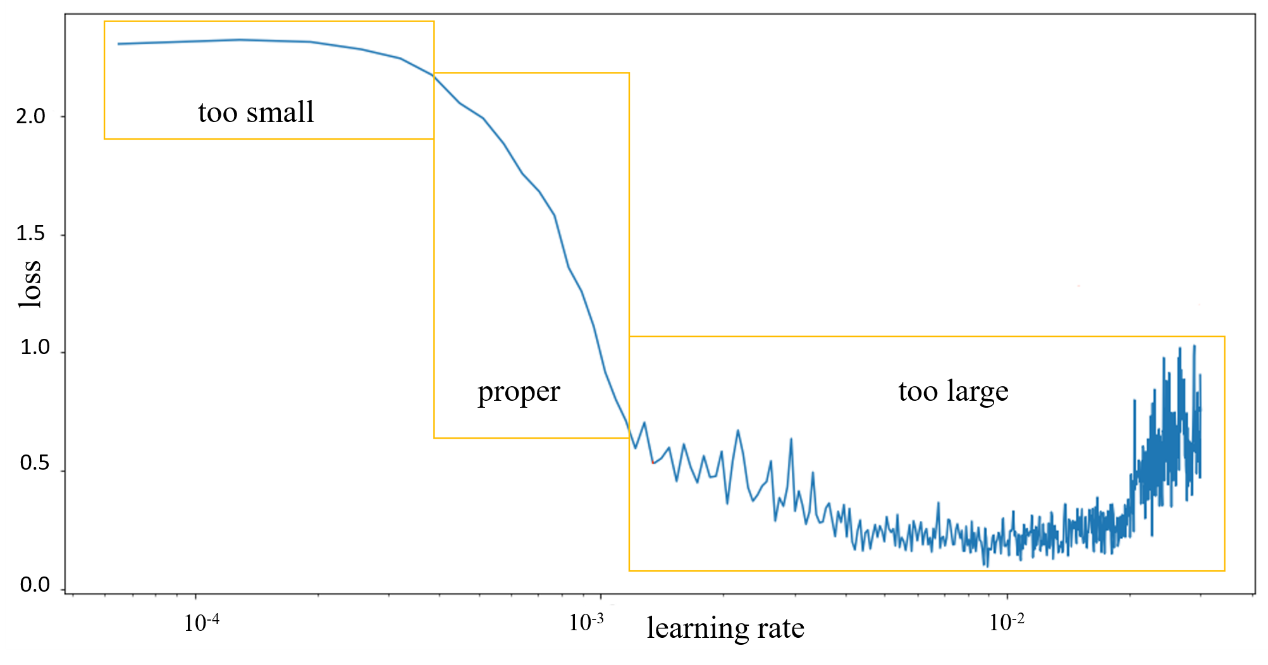}
    \caption{Effect of learning rate (adapted figure from: https://www.jeremyjordan.me/nn-learning-rate/)}
    \label{Figure 4}
\end{figure}

\indent
LR is a positive scalar that determines the length of step during SGD~\citep{goodfellow16}. In most cases, the LR must be manually adjusted during model training, and this adjustment is often necessary for enhanced accuracy~\citep{bengio12}.An alternative choice of fixed LR is a varied LR over the training process. This method is referred to as the LR schedule or LR decay~\citep{goodfellow16}. Adaptive LR can be adjusted in response to the performance or structure of the model, and it is supported by a learning algorithm~\citep{smith17,brownlee19}.\\
\indent
Constant LR, as the simplest schedule, is often set as the default by deep learning frameworks (e.g., Keras). It is an important but tricky task to determine a good value for LR. With a proper constant LR, the network is able to be trained to a passable but unsatisfactory accuracy because the initial value could always be overlarge, especially in the final few steps. A small improvement based on the constant value is to set an initial rate as 0.1, for example, adjusting to 0.01 when the accuracy is saturated, and to 0.001 if necessary~\citep{tanksale18}.\\
\indent
Linear LR decay is a common choice for researchers who wish to set a schedule. It changes gradually during the training process based on time or step. The basic mathematical function of linear decay is
\[
lr \;=\;\frac{lr_0}{1+kt}
\]
where \emph{lr$_0$} and \emph{k} are hyper-parameters for linear learning decay, indicating the initial LR and decay rate. A typical choice of \emph{t} could be training time or number for self-iteration. If \emph{t} is training time, the schedule of LR is a continuous change, as shown in Figure \ref{Figure 1}. If t stands for the number of iterations, the LR drops every few iterations. A typical choice is to drop the LR by half every 10 epochs~\citep{lau2017learning} by 0.1 every 20 epochs~\citep{li15}. This has the following mathematical form:
\[
lr \;=\;\frac{lr_0}{1+deacay*selfiteration}
\]
\indent
Compared with LR itself, the deep learning model is less sensitive to \emph{lr$_0$} and \emph{k}.\\
\indent
Exponential decay is another widely used schedule~\citep{li15}.Compared with linear decay, an exponential schedule provides a more drastic decay at the beginning and a gentle decay when approaching convergence (Figure \ref{Figure 2}). The basic mathematical form of exponential decay is as follows:
\[
lr \;=\;{lr_0}\cdot exp(-kt)
\]
where \emph{lr$_0$} and \emph{k} are hyper-parameters for exponential learning decay. Similar to linear decay, \emph{t} could be the number of self-iterations or epochs. If the number of epochs is used, the LR can be expressed with a similar form:
\[
lr \;=\;{lr_0}\cdot floor^{\frac{epoch}{epochsdrop}}
\]
\indent
In addition to the initial LR and epoch drop rate, \emph{floor} is another hyper-parameter. Generally, the LR is set to drop every 10 epochs (\emph{epochsdrop}=10) by half (\emph{floor}=0.5),but these can be decided by the model. EfficientNet~\citep{tan19b} uses an exponential schedule when trained with ImageNet~\citep{krizhevsky12}. Its choice for \emph{lr$_0$}, \emph{floor} and \emph{epochsdrop} are too tricky to be designed purely by hand. For example, the LR is updated every 2.4 epochs (\emph{epochsdrop}=2.4) by 0.97 (\emph{floor}=0.97). Automatic hyper-parameter tuning is supposed to be applied in this case.\\
\indent
A typical LR schedule may encounter some challenges when applied to a certain model. One problem is that users must determine all hyper-parameters in the schedule in advance of training, which is a task that depends on experience. Another problem is that in the abovementioned schedules, the LR only varies with time or steps, and the same value is applied for all layers in a model. The first problem is typically solved with automated HPO or the cyclical LR method~\citep{smith17}. The LR is updated in a triangle rule within a certain bound value, and the bound value is decayed in a certain cyclic schedule (Figure \ref{Figure 3}).\\
\indent
Regarding the second problem, this study~\citep{you17} suggested an algorithm based on layer-wise adaptive rate scaling (LARS). In their study, every layer had a local LR related to its weight and gradient. Users decide how to adjust the LR according to the layers and related hyper-parameters.\\
\indent
In reality, selecting the optimal LR or its optimum schedule is a challenge. A small LR leads to slow convergence, whereas a large LR may prevent the model from converging (Figure \ref{Figure 4}). The LR schedule must be adjusted according to the optimizer algorithm~\citep{lau2017learning}. If the LR schedule must be determined in advance, it is recommended to simultaneously tune the hyper-parameters of the LR schedule and corresponding optimizers.\\
\indent
The choice of LR varies with specific tasks, but generally some tips exist that are based on our experience. They can be viewed as a general rule for hyper-parameter tuning.\\
\begin{itemize}
\item[-]In practice, it is difficult to decide the importance of a hyper-parameter if one has no experience of it. A sensitivity test is suggested~\citep{hamby94,breierova1996introduction} to ensure the influence of a certain hyper-parameter.
\item[-]Initial values of hyper-parameters are influential and must be carefully determined. The initial LR could be a comparatively large value because it will decay during training. In the early stage of training, a large LR will lead to fast convergence with fewer risks (Figure \ref{Figure 4})
\item[-]Use log scale to update the LR. Thus, exponential decay could be a better choice. An exponential schedule could be applicable for many other tuning hyper-parameters, such as momentum and weight decay.
\item[-]Try more schedules. Exponential decay is not always the best choice; it depends on the model and dataset.
\end{itemize}

\subsection{Optimizer}
\begin{figure}
    \centering
    \includegraphics{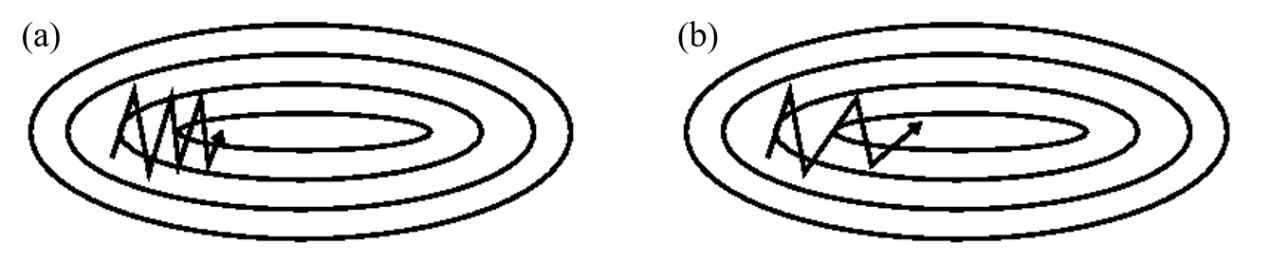}
    \caption{(a) SGD without momentum; (b) SGD with momentum (Source: https://www.willamette.edu/~gorr/classes/cs449/momrate.html)}
    \label{Figure 5}
\end{figure}

\indent
Optimizers, or optimization algorithms, play a critical role in improving accuracy and training speed. Hyper-parameters related to optimizers include the choice of optimizer, mini-batch size, momentum, and beta. Selecting an appropriate optimizer is a tricky task. This section discusses the most widely adopted optimizers (mini-batch gradient descent, RMSprop, and Adam), related hyper-parameters, and suggested values.\\
\indent
The aim of mini-batch gradient descent~\citep{li14}is to solve the following two problems. Compared with vanilla gradient descent~\citep{li15},mini-batch gradient descent (mini-batch GD) accelerates the training process, especially on a large dataset. Compared with SGD with a mini-batch size of 1, mini-batch GD reduces the noise and increases the probability of convergence~\citep{ng2017improving}.Mini-batch size is a hyper-parameter, and the value is highly related to the memory of the computation unit. Its value is suggested to be a power of 2 because of the access of CPU/GPU memory. The model runs faster if a power of 2 is used as the mini-batch size, and 32 could be a good default value~\citep{bengio12,masters18}. The maximum mini-batch size fitting in CPU/GPU memory. With a constant LR, researchers may find that their model oscillates within a small range but does not exactly converge in the last few steps (Figure \ref{Figure 4}). An opinion exists that for enhanced accuracy, mini-batch size and LR could be re-optimized after other hyper-parameters are fixed~\citep{masters18}. Vanilla mini-batch GD without momentum may take longer than some more recent optimizers, and the convergence relies on a robust initial value and LR. As the foundation of more advanced optimization algorithms, mini-batch GD is still used but not widely in recent publications~\citep{ruder16}.\\
\indent
An improved method for solving the problem of oscillation and convergence speed is SGD with momentum~\citep{loizou17}. The momentum method~\citep{qian99} accelerates the standard SGD by calculating the exponentially weighted averages for gradients (Figure \ref{Figure 5}). Furthermore, it helps the cost function go in the correct direction through adding a fraction beta of the update vector:
\[
{v_{dw}} \;=\;\beta{v_{dw}}+(1-\beta)dw
\]
\[
w \;=\;w-lr*{v_{dw}}
\]
where \emph{w} indicates the weight. Here, momentum $\beta$  is also a hyper-parameter. The momentum term is usually set as 0.9, or 0.99 or 0.999 if necessary. This reduces the oscillation by strengthening the update in the same direction and decreasing the change in different directions~\citep{sutskever13}.\\
\indent
Root mean square prop (RMSprop) is one of the most widely used optimizers in the training of deep neural networks~\citep{karpathy16}. RMSprop accelerates the gradient descent in a manner similar to Adagrad and Adadelta~\citep{zeiler12}, but it exhibits superior performance when steps become smaller. It can be considered the development of Rprop~\citep{igel00} for mini-batch weight update. In RMSprop, the LR is adapted by dividing by the root of the squared gradient. Compared with the original GD, RMSprop slows the vertical oscillation and accelerates the horizontal movement from a much larger LR~\citep{bushaev18}, and it is implemented as follows. RMSprop uses exponential weighted averages of the squares instead of directly using \emph{dw} and \emph{db}:\\
\[
{S_{dw}} \;=\;\beta{S_{dw}}+(1-\beta)dw^2
\]
\[
{S_{db}} \;=\;\beta{S_{db}}+(1-\beta)db^2
\]
\[
w \;=\;w-lr*\frac{dw}{sqrt(S_{dw})}
\]
where $\beta$ is the moving average parameter, suggested as 0.9~\citep{hinton12b}. \emph{$S_{dw}$} is relatively small and thus \emph{dw} is divided by a relatively small number, whereas \emph{$S_{db}$} is relatively large and thus \emph{db} s divided by a larger number. In this way, RMSprop endures a slower movement in the bias direction (vertical) and a faster move in the weight direction (horizontal; Figure 6).\\
\begin{figure}
    \centering
    \includegraphics{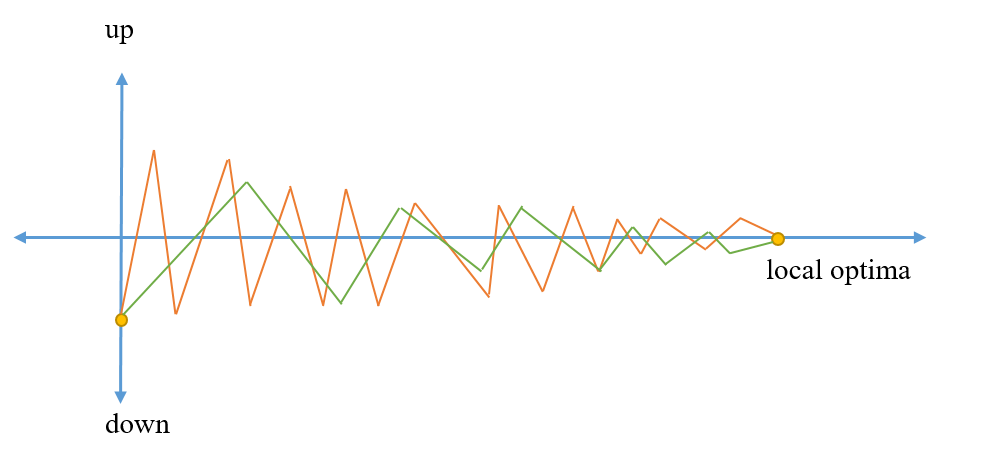}
    \caption{Comparison of SGD (orange) and RMSprop (green) for optimization}
    \label{Figure 6}
\end{figure}
\indent
Adaptive momentum estimation (Adam) also achieves good results quickly on most neural network architectures. As a combination of GD with momentum and RMSprop, Adam adds bias-correction and momentum to RMSprop, enabling it to slightly outperform RMSprop in the late stage of optimization. It calculates the exponential weighted average as well as squares of past gradients. Parameters \emph{$\beta_1$} and \emph{$\beta_2$} control the decay rate as shown below. Moreover, Adam retains an exponential decaying average of past gradients. Its requirement for memory is in the middle of GD with momentum and RMSprop. In practice, Adam is suggested as the default optimization algorithm for deep learning training~\citep{karpathy16}. It has more hyper-parameters compared with other optimizers, but works well with little tuning of hyper-parameters except for LR.\\
\indent
\emph{$v_{dw}$} and \emph{$v_{db}$} are updated as with GD with momentum:\\
\[
{v_{dw}} \;=\;{\beta_1}{v_{dw}}+(1-{\beta_1})dw
\]
\[
{v_{db}} \;=\;{\beta_1}{v_{db}}+(1-{\beta_1})db
\]
\indent
\emph{$S_{dw}$} and \emph{$S_{db}$} are updated as with RMSprop:\\
\[
{S_{dw}} \;=\;{\beta_2}{S_{dw}}+(1-{\beta_2})dw^2
\]
\[
{S_{db}} \;=\;{\beta_2}{S_{db}}+(1-{\beta_2})db^2
\]
\indent
The biases are corrected as follows:
\[
{v_{dw}^{corrected}} \;=\;{v_{dw}}+(1-{\beta_1^t})
\]
\[
{v_{db}^{corrected}} \;=\;{v_{db}}+(1-{\beta_1^t})
\]
\[
{S_{dw}^{corrected}} \;=\;{S_{dw}}+(1-{\beta_w^t})
\]
\indent
weights \emph{w} and biases \emph{b} are updated as follows:\\
\[
w \;=\;w-lr*\frac{v_{dw}^{corrected}}{sqrt(S_{dw}^{corrected}+\epsilon)}
\]
\[
b \;=\;b-lr*\frac{v_{db}^{corrected}}{sqrt(S_{db}^{corrected}+\epsilon)}
\]
\indent
For the adjustment of LR, the previous section can be referred to, and it is reasonable to start from 0.001. \emph{$\beta_1$} is the parameter of momentum, with a suggested value of 0.9; \emph{$\beta_2$} is the parameter of RMSprop, with a suggested value of 0.999; $\epsilon$ here is used to avoid dividing by 0; and $10^{-8}$ is the default value. The default configurations are suitable for most problems, and therefore Adam is relatively easy to apply~\citep{brownlee17}. The suggested values are adapted by most deep learning frameworks, including TensorFlow~\citep{abadi16}, Keras~\citep{chollet18}, Caffe~\citep{jia14}, MxNet~\citep{chen15}, and PyTorch~\citep{paszke19}.\\

\indent
Generally, optimization algorithms must be adjusted along with the LR schedule. Most hyper-parameters for neural network training are highly related to optimizers and LR. RMSprop and Adam can be applied in similar situations. In most cases, Adam can act as the default optimizer~\citep{ruder16} because it is a more reliable improvement based on RMSprop. If Adam is used as the optimizer, LR must be adjusted accordingly, whereas default values are sufficient for most other hyper-parameters. Compared with RMSprop and Adam, SGD with momentum may require more time to find the optimum.\\
\indent
The abovementioned subsections discussed the most critical hyper-parameters related to model training. According to lecture notes of Ng (2017), the importance of hyper-parameters are listed as follows:\\
\begin{enumerate}
\item[-] LR, as in our discussion above.
\item[-] Momentum beta, for RMSprop etc.
\item[-] Mini-batch size, as in our discussion above, which is critical for all mini-batch GD.
\item[-] Number of hidden layers, which is a hyper-parameter related to model structure.
\item[-] LR decay, as in our discussion above.
\item[-] Regularization lambda, which is used to reduce variance and avoid overfitting.
\item[-] Activation functions, which is used to add nonlinear elements.
\item[-] Adam \emph{$\beta_1$}, \emph{$\beta_2$} and \emph{$\epsilon$}, as in our discussion above.
\end{enumerate}
We can notice that hyper-parameters included in the list are not only related to training, but also determine the structure of model. , And therefore in the next section, we only talk about the number of hidden layers,  regularization lambda, and activation function.\\

\subsection{Model Design-Related Hyper-Parameters}
The number of hidden layers(\emph{d}) is a critical parameter for determining the overall structure of neural networks, which has a direct influence on the final output~\citep{hinton06}. Deep learning networks with more layers are likely to obtain more complex features and relatively higher accuracy. Upscaling the neural network by adding more layers is a regular method of achieving better results. In this way, a baseline structure is repeated to increase the receptive field. For example, users can choose ResNet-18 to ResNet-200 according to their need for accuracy~\citep{he16}. This study~\citep{tan19b} created a series of neural networks with sets of depth and width hyper-parameters, obtaining high accuracy with low computational resources and a limited number of parameters.\\
\indent
The number of neurons in each layer(\emph{w}) must also be carefully considered. Too few neurons in the hidden layers may cause underfitting because the model lacks complexity~\citep{oppermann19}. By contrast, too many neurons may result in overfitting and increase the training time. The suggestions~\citep{heaton17} could be a good start for tuning the number of neurons:\\
\[
	w_{input} \;<\; w \;<\; w_{output}
\]
\[
	w \;=\; \frac{2}{3}w_{input}+ w_{output}
\]
\[
	w \;<\; 2w_{output}
\]
where \emph{$w_{input}$} is the number of neurons for the input layer, and \emph{$w_{output}$} is the number of neurons for the output layer.\\
\indent
In contrast to adding layers or neurons in each layer, regularization is applied to reduce the complexity of neural network models, especially those with insufficient training data. Overfitting usually occurs in complex neural networks with many layers and neurons, and the complexity can be reduced using regularization methods. A regularization term is added to avoid overfitting and address feature selections. The general expression of the regularization term is as follows~\citep{ng04}:\\
\[
	cost \;=\; loss function + regularization term
\]
\indent
The L1 term uses the sum of absolute values of weights. The regression model used for L1 regularization is also called \emph{Lasso Regression} or L1 norm:
\[
    w* \;=\; \sum_{i=0}^N (y_i-\sum_{j=0}^M {x_ij}{W_j})^2 + {\lambda}\sum_{j=0}^M {\vert{W_j}\vert}
\]
\indent
The L2 term uses the sum of least square values of weights. The regression model used for L2 regularization is also called Ridge Regression or L2 norm:\\
\[
    w* \;=\; \sum_{i=0}^N (y_i-\sum_{j=0}^M {x_ij}{W_j})^2 + {\lambda}\sum_{j=0}^M {W_j^2}
\]

where \emph{$\lambda$} is the regularization hyper-parameter. An overlarge \emph{$\lambda$} will oversimplify the structure of the deep learning network because some weights are close to zero, whereas an undersized \emph{$\lambda$} is not strong enough to reduce weights. In practice, L2 regularization is more widely used, mainly because of its computational efficiency; however, L1 regularization prevails over L2 on sparse properties~\citep{yan16}. Models generated using L1 norm are simpler and interpretable, while L2 provides superior predictions. L1 and L2 have pros and cons, as shown in Table \ref{Table 1} adapted from~\citep{khandelwal18}.\\
\begin{table}
    \centering
    \begin{tabular}{c|c}
     \toprule
     L1 & L2\\
     \midrule
     {Penalizes the sum of the absolute value of weights} & {Penalizes the sum of square weights}\\
     {Sparse solution} & {Nonsparse solution}\\
     {Multiple solution} & {One solution}\\
     {Built-in feature selection} & {No feature selection}\\
     {Robust to outliers} & {Not robust to outliers}\\
     {Unable to learn complex patterns} & {Able to learn complex data patterns}\\
     \bottomrule
    \end{tabular}
    \caption{Comparison between L1 and L2}
    \label{Table 1}
\end{table}
\indent
In addition to the abovementioned methods, data augmentation~\citep{wang17} and dropout~\citep{srivastava14} are commonly used regularization techniques. Data augmentation is the most straightforward, especially for image classification and object detection. The data augmentation method creates and adds fake data to the training dataset to avoid overfitting. Operations including image transformation, cropping, flipping, color adjustment, and image rotation can improve the generalization in different cases~\citep{devries17}. For documentation recognition~\citep{lecun98}, several transformations for enhanced model accuracy and robustness are mainly applied. For character recoginition task~\citep{bengio11} image transformations and occlusions for their handwritten characters were combined.\\
\begin{figure}
    \centering
    \includegraphics{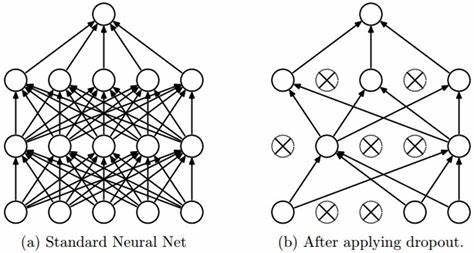}
    \caption{Comparison of a neural network before and after dropout is applied~\citep{srivastava14} }
    \label{Figure 7}
\end{figure}
\indent
Dropout is a technique used to select neurons randomly with a given probability that are not used during training, which makes the network less sensitive to the specific weights of neurons~\citep{hinton12a}. As a result, the simplified version of a neuron network can reduce overfitting (Figure  \ref{Figure 7}). The probability of neuron deactivation applied for each weight-updated step is suggested to be around 20\%-50\%. An overlarge dropout rate will over-simplify the model, whereas a small value will have little effect. In addition, a larger LR with decay and a larger momentum are suggested because fewer neurons updated with dropout requires more update for each batch~\citep{brownlee16}.\\

\begin{figure}
    \centering
    \includegraphics[width=13.5cm,height=8.5cm]{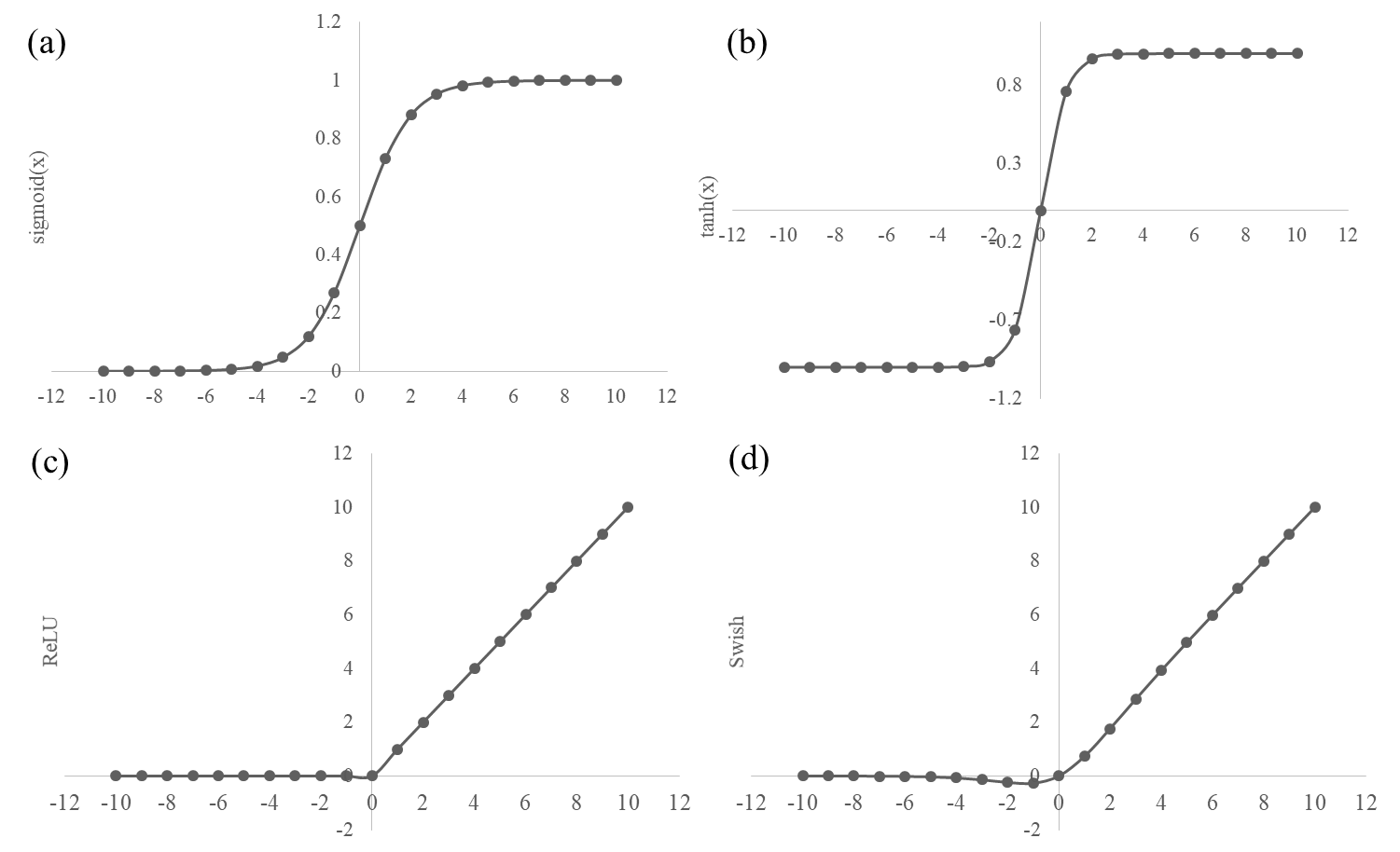}
    \caption{Comparison of activation functions: (a) sigmoid; (b) tanh; (c) ReLU; and (d) Swish }
    \label{Figure 8}
\end{figure}

\indent
Activation functions are crucial in deep learning for introducing nonlinear properties to the output of neurons. Without an activation function, a neural network will simply be a linear regression model that is unable to represent complicated features of data. Activation functions should be differentiable to compute the gradients of weight and perform backpropagation\citep{walia17}. The most popular and widely used activation functions include sigmoid, hyperbolic tangent (tanh), rectified linear units (ReLU)~\citep{nair10}, Maxout~\citep{goodfellow13}, and Swish~\citep{ramachandran17a}. Automatic search techniques have been applied in search of proper activation functions, including the structure of functions and related hyper-parameters~\citep{ramachandran17b}.\\
\indent
Sigmoid functions are the most frequently used activation layer with a form of $f(x)\;=\; \frac{1}{1+e^{-x}}$. The output of a sigmoid function is always between 0 and 1 (Figure \ref{Figure 8}a) when negative infinite to positive infinite are encountered, and thus it will not blow up the activation. They provide nonlinearity and make clear distinctions, but in the past few years they have fallen out of popularity because of some major disadvantages. With sigmoid function, one finds that the output f(x) responds very little to the input x, which gives rise to the vanishing gradient problem~\citep{wang2019vanishing}. This may not be a big problem for networks with a few layers, but it causes weights that are too small when more layers are required in a network. Another problem is it not being zero-centered, which means the average of data lies on the zero. During backpropagation, weights will either be all positive or all negative, making an undesirable zigzag on weights and making optimization more difficult. Softmax is a function similar to a sigmoid function with each input within (0, 1). The major difference is that a sigmoid function is applied for binary classification, whereas a softmax function is for a multinomial classification. With j classes, the predicted probability of $i^{th}$ class is as $f(x_i)\;=\; \frac{exp(x_i)}{\sum jexp(x_i)}$. In addition, softmax function allows a probabilistic interpretation. With the softmax function, a high input value results in a higher probability than others, and the sum of probabilities will be 1. Whereas with the sigmoid function, a high value will lead to the high probability as well, but not the higher probability, and the sum of probabilities is not necessarily to be 1. While creating neural networks, sigmoid function is used as activation function, and softmax function is usually used in output layers.\\
\indent
Hyperbolic tangent (tanh) is in a form of $f(x)\;=\; \frac{1-e^{-2x}}{1+e^{-2x}}$. It is zero-centered and the output ranges between minus one and one (Figure \ref{Figure 8}b). It is in a similar structure to a Sigmoid function but with a steeper derivative. Tanh is more efficient than a Sigmoid function and has a wider range for faster learning and grading~\citep{kizrak19}, but it still has the problem of vanishing gradients. Sigmoid and tanh cannot be applied for networks with many layers.\\

\indent
In recent years, ReLU has become the most widely used activation function~\citep{agarap18}. More advanced activation functions have arisen but ReLU and its variants (e.g., leaky ReLU, PReLU, ElU, and SeLU) are still the most typical and popular methods (Figure \ref{Figure 8}c).\\
\[
f(x)\;=\;\left\{
        \begin{array}{rcl}
        0, & x\;\leq\;0\\
        x, & x\;> \;0\\
        \end{array}
        \right.
\]
\indent
ReLU is mathematically simple and solves the problem of vanishing gradients. Compared with sigmoid and tanh, ReLU does not covert almost all neurons to be activated in a similar way, and it is great for multi-layer networks. However, ReLU has two major problems: the first is that its range is from zero to infinite, which means that it can blow up the activation, and the second is its sparsity. For a negative x, ReLU yields some weights to 0 and will never activate at any data point again~\citep{lu19}. To solve the second problem, leaky ReLU was proposed, which uses a very small slope (e.g., 0.01 x) for a negative x to keep the update alive. Despite these problems, ReLU is suggested to be used by researchers who have little knowledge about their tasks or activation functions because it involves highly simple mathematical operations. For deep learning networks, ReLU could be the default choice in terms of ease and speed. If the problem of vanishing gradients occurs, leaky ReLU could be the alternative.\\
\indent
Maxout activation is a generalization of ReLU and leaky ReLU, and is defined as $f_i(x)\;=\;max z_{ij},z_{ij}\;=\;x^TW_{...ij} + b_{ij}, W\in\mathbf{R^{d\times m\times k}}$. It is actually a layer where the activation is the max of input. Maxout activation is a learnable activation function fixing all problems with ReLU, and it often works in combination with dropout. However, it doubles the total number of parameters because each neuron has an addition weight and bias.\\
\indent
Swish was proposed by Google Brain as a competitor of ReLU. It is defined as $f(x)\;=\;x\cdot sigmoid(x)\;=\;\frac{x}{1+e^{-x}}$ (Figure \ref{Figure 8}d), the multiplication of input x and sigmoid function of x. Swish has been claimed to outperform ReLU with similar simple calculations, but it is not as popular as ReLU. It has similar shape to ReLU, with bounded below and unbounded above. For a negative large x, Swish yields it into zero, whereas the values for a small negative x are captured. In addition, Swish is a smooth curve and thus it is more favorable if a smooth landscape is required.\\

\section{Search Algorithms and Trial Schedulers on Hyper-Parameter Optimization}
HPO is a problem with a long history, but it has recently become popular for the application of deep learning networks. The general philosophy remains the same: determine which hyper-parameters to tune and their search space, adjust them from coarse to fine, evaluate the performance of the model with different parameter sets, and determine the optimal combination~\citep{strauss07}. This present study discusses the optimization methods for the second and third step: how to adjust hyper-parameters and how to evaluate the model’s performance.\\
\indent
For deep learning networks, the choice of hyper-parameters influences both the structure of the network (e.g., the number of hidden layers) and training accuracy (e.g., LR). The hyper-parameters could be integers, floating points, categorical data, or binary data, with a distribution of the search space. Mathematically, HPO is a process of finding a set of hyper-parameters to achieve minimum loss or maximum accuracy of an objective network. The objective function is described as follows~\citep{feurer19}:\\
\[
    \lambda^* \;=\; argmin\;or\; argmax E_{(D_{train}, D_{valid})\sim D}V(L, A_{\lambda}, D_{train}, D_{valid})
\]
\indent
To maximize or minimize the V function using algorithm \emph{$A_{\lambda}$} applied for hyper-parameters, the model is trained with dataset \emph{$D_{train}$} and validated with dataset \emph{$D_{valid}$}.\\
\indent
The state-of-the-art algorithms for HPO can be classified into two categories: search algorithms and trial schedulers. In general, search algorithms are applied for sampling whereas trial schedulers mainly deal with the early stopping methods for model evaluation. This section first discusses popular search algorithms followed by state-of-the-art early stopping strategies.\\

\subsection{Search Algorithms}
\subsubsection{Grid Search}
\begin{figure}
    \centering
    \includegraphics{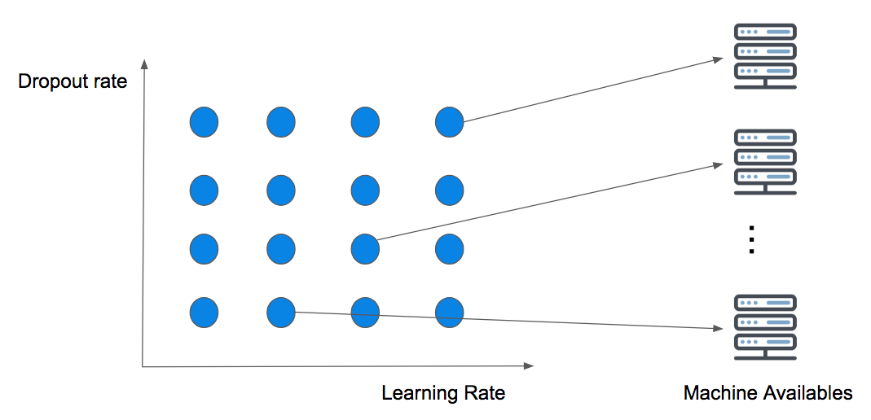}
    \caption{Grid search on dropout rate and learning rate in a parallel concurrent execution~\citep{harrington18}  }
    \label{Figure 9}
\end{figure}

Grid search is a basic method for HPO. It performs an exhaustive search on the hyper-parameter set specified by users. Users must have some preliminary knowledge on these hyper-parameters because it is they who generate all candidates. Grid search is applicable for several hyper-parameters with limited search space.\\
\indent
Grid search is the most straightforward search algorithm that leads to the most “accurate” predictions—as long as sufficient resources are given, the user can always find the optimal combination~\citep{joseph18}. It is easy to run grid search in parallel because every trial runs individually without the influence of time sequence. Results for one trial are independent of those from other trials. Computational resources can be allocated in a highly flexible manner (Figure \ref{Figure 9}). However, grid search suffers from the curse of dimensionality because the consumption of computational resources increases exponentially when more hyper-parameters are awaiting tuning. Of course, methods exist for dimensionality reduction (e.g., PCA), but any method comes with a cost on the robustness of the model~\citep{yiu19}. In addition, a limited sampling range is acceptable for grid search because too many configurations are not desirable. In practice, grid search is almost only preferable when users have enough experience of these hyper-parameters to enable the definition of a narrow search space and no more than three hyper-parameters need to be tuned simultaneously.\\
\indent
Although other search algorithms may have more favorable features, grid search is still the most widely used method because of its mathematical simplicity~\citep{bergstra2012random}.\\

\subsubsection{Random Search}
\begin{figure}
    \centering
    \includegraphics[width=10.5cm,height=5.3cm]{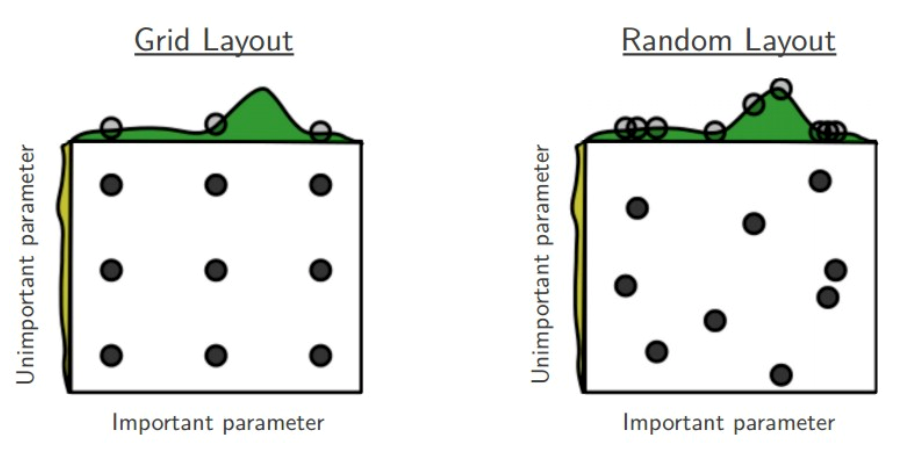}
    \caption{Layout comparison between grid search and random search~\citep{bergstra2012random}  }
    \label{Figure 10}
\end{figure}

Random search~\citep{bergstra2012random} is a basic improvement on grid search. It indicates a randomized search over hyper-parameters from certain distributions over possible parameter values. The searching process continues till the predetermined budget is exhausted, or until the desired accuracy is reached. Random search is similar to grid search but has been proven to create better results because of the following two benefits~\citep{maladkar18}: first, a budget can be assigned independently according to the distribution of search space, whereas in grid search the budget for each hyper-parameter set is a fixed value \emph{$B^{1/N}$}, Where \emph{B} is the total budget and \emph{N} is the number of hyper-parameters. Therefore, random search may perform better especially when some hyper-parameters are not uniformly distributed. In this search pattern, random search is comparatively more likely to find the optimal configuration than grid search (Figure \ref{Figure 10}). Second, although obtaining the optimum using random search is not promising, it is quite certain that greater time consumption will lead to a larger probability of finding the best hyper-parameter set. This logic is known as Monte Carlo techniques, which is popular when deal with large volume datasets in multi-dimensional deep learning scenarios~\citep{harrison10}. By contrast, for grid search, a longer search time cannot guarantee better results. Easy parallelization and flexible resource allocation are also dominant advantages of random search~\citep{krivulin05}. \\
\indent
In most cases, random search is more effective than grid search, but it is still a computationally intensive method. The use of random search is suggested in the early stage of HPO to rapidly narrow down the search space, before using a guided algorithm to obtain a finer result (from a coarse to fine sampling scheme)~\citep{ng2017improving}. Random search is often applied as the baseline of HPO to measure the efficiency of newly designed algorithms. Random search generally takes more time and computational resources than other guided search methods.\\

\subsubsection{Bayesian Optimization and Its Variants}
Bayesian optimization (BO) is a traditional algorithm with decades of history. It was raised by Mockus~\citep{mockus75,mockus78}, and later became popular when it was applied to the global optimization problem ~\citep{jones98}. BO is a typical method for almost all types of global optimization, and is aimed at becoming “less wrong” with more data~\citep{Koehrsen18}. It is a sequential model-based method aimed at finding the global optimum with the minimum number of trials. It balances exploration and exploitation~\citep{silver14} to avoid trapping into the local optimum. Exploitation is the process of making the best decision based on current information (Figure 11, right), whereas with exploration, the model will collect more information (Figure 11, left). BO outperforms random search and grid search in two aspects: the first is that users are not required to possess preliminary knowledge of the distribution of hyper-parameters, and the second is posterior probability, which is the core idea of BO. The process of BO is described as follows: a probability surrogate model of objectives is built, and then every attempt is made based on the assessment of previous trials, and furthermore, the probability model is updated for the next trial until the most promising hyper-parameter set is chosen. Compared with grid search and random search, BO is more computationally efficient with fewer attempts required to find the best hyper-parameter set, in particular, especially when very costly objective functions are encountered. BO has another remarkable advantage over grid search and random search—it is applicable regardless of whether the objective function is stochastic or discrete, or convex or nonconvex.\\
\begin{figure}
    \centering
    \includegraphics[width=14.5cm,height=4cm]{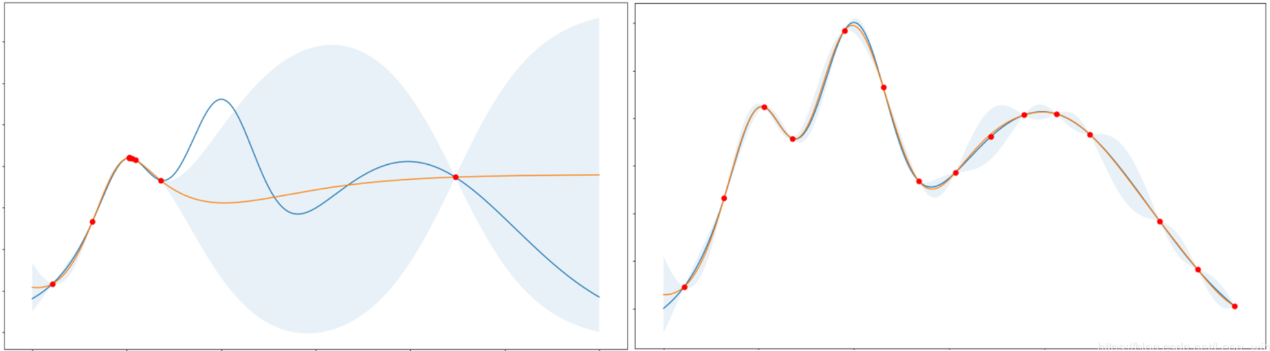}
    \caption{Exploration-oriented (left) and exploitation-oriented Bayesian optimization (right); the shade indicates uncertainty. }
    \label{Figure 11}
\end{figure}
\indent
This subsection presents a basic overview of the Bayesian algorithm, describes the variants based on BO, and discusses its recent applications in deep learning.\\
\indent
BO consists of two key ingredients: a Bayesian probability surrogate model to model the objective function, and an acquisition function to determine the next sampling point. The algorithm’s process can be described as follows: (1) build a prior distribution of the surrogate model; (2) obtain the hyper-parameter set that performs best on the surrogate model; (3) compute the acquisition function with the current surrogate model; (4) apply the hyper-parameter set to the objective function; and (5) update the surrogate model with new results. Steps 2–5 are repeated until the optimal configuration is found or the resource limit is reached~\citep{brecque18}.\\
\indent
The first ingredient – the surrogate model – could be a specific analytic function or nonparametric model. The prior distribution describes the hypothesis on objective function, and the posterior distribution is fitted based on observational data thus far. The second ingredient – the acquisition function – plays a role in balancing exploration and exploitation. It ideally minimizes the loss function to select the optimal candidate points. Many probability models could be used in the BO algorithm~\citep{eggensperger13}, but Gaussian process (GP)~\citep{rasmussen03} is overwhelmingly the most widely used. Various acquisition functions have been proposed in previous studies, including the probability of improvement~\citep{kushner64}, GP upper confidence bound (GP-UCB)~\citep{lai85,srinivas09}, predictive entropy search~\citep{hernandez14}, and even a portfolio containing multiple acquisition strategies~\citep{hoffman11}. Among them, the expected improvement algorithm~\citep{mockus78} is definitely the most common choice. In this study, we only discussed the most popular choice of surrogate model and acquisition function. A lack of space limits further discussion on this topic; please refer to the review ~\citep{shahriari15} for more information.\\
\begin{figure}
    \centering
    \includegraphics[width=14.5cm,height=12.7cm]{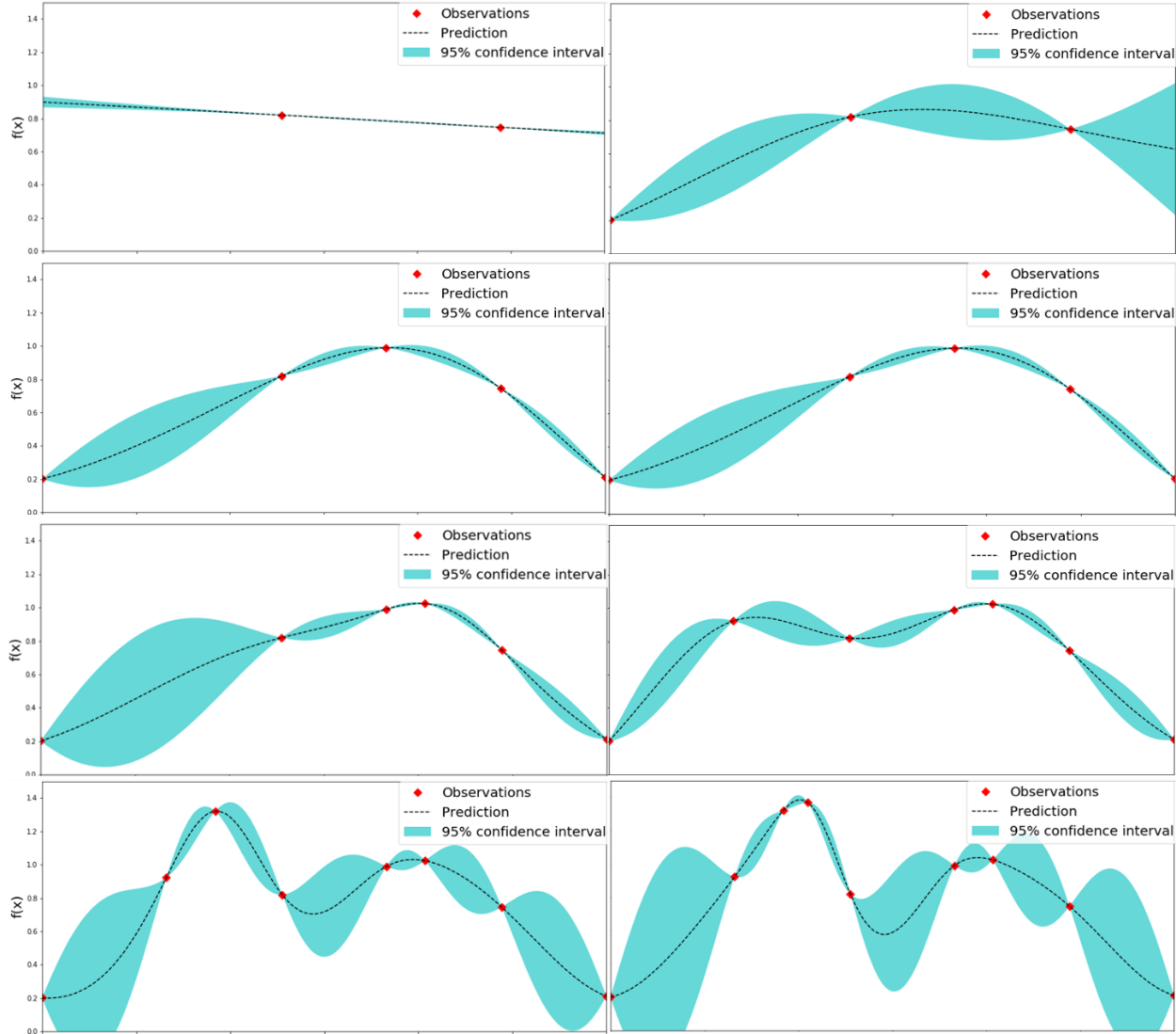}
    \caption{Gaussian process with 2–9 samplings }
    \label{Figure 12}
\end{figure}
\indent
In BO, GP could be the default choice of surrogate model for objective functions (Figure \ref{Figure 12}). It is popular for two reasons: first, as a nonparametric Bayesian statistical method, GP is a method of obtaining flexible models. For a nonparametric model, the number of parameters is decided by the dataset size, without the requirement for it to be determined in advance. Second, GP takes multivariate Gaussian distribution as the prior distribution to infinite numbers of real-valued variables. In this stochastic process, any finite subset for random variables ${x_1,..., x_n}\in\chi $ follows multivariate Gaussian distribution~\citep{do07}:\\
\[
    f(\lambda)\;\sim\; GP\Big(\mu(\lambda),k(\lambda,\lambda')\Big)
\]
where $\mu(\cdot)$ is the mean vector,\\
\[
    \mu(\lambda)\;=\;E[x]\;=\;
    \left[ \begin{array}{c}
    \mu(x_1)\\
    \vdots\\
    \mu(x_n)
    \end{array}\right]
\]
and $k(\cdot,\cdot')$ is the covariance function,\\
\[
    k(\lambda, \lambda')\;=\;E\Big[\Big(x-\mu(x)\Big)\Big(x'-\mu(x')\Big)\Big]\;=\;
    \left[ \begin{array}{ccc}
    k(x_1,x_1) & \ldots &  k(x_1,x_m)\\
    \vdots & \ddots & \vdots\\
    k(x_m,x_1) & \ldots &  k(x_m,x_m)
    \end{array}\right]
\]
for any $x,x'\in X$. Similar to a Gaussian distribution defined by mean and covariance, a Gaussian process is a high-dimensional multivariate Gaussian defined with the mean vector and covariance matrix. The mean function controls the smoothness and amplitude of samples while the choice of covariance function or kernel function \emph{$k(\lambda, \lambda')$} determines the quality of the surrogate model~\citep{rasmussen03}. The most commonly used mean function is $\mu(\lambda)=0$, and the most popular kernel is the squared exponential kernel or Gaussian kernel $k_SE(x,x')=exp(-\frac{{\Vert x \Vert}^2}{2l^2})$. Mat\'ern 5/2 kernel is widely used because of its flexibility~\citep{duvenaud14}  with learnable length scale parameters are common default kernel functions for many applications (e.g., Spearmint and MOE). Kernel function can be combined in various ways, and thus two or more kernels can be used simultaneously for GP.\\
\indent
Prediction following a normal distribution $p(y|x,D)= N(y|\hat\mu,{\hat\sigma}^2)$~\citep{fasshauer15} with $\mu_0(x)=0$ has a posterior distribution of\\
\[
    \mu_n(x)\;=\;k(x)^T(K+{\sigma}^2I)^{-1}y
\]
\[
    \sigma_n^2(x)\;=\;k(x,x)-k(x)^T(K+{\sigma}^2I)^{-1}k(x)
\]
When the surrogate model is chosen, posterior distribution at any point could be determined accordingly by the mean function and kernel function. Mean indicates the expected results: a larger mean value implies a higher possibility of the optimum option. Kernel indicates the uncertainty: a larger covariance value implies that exploration is worthwhile. To avoid becoming trapped into the local optimum, whether choosing the location of the next sampling point with a larger mean or larger uncertainty is determined using acquisition functions. \\
\indent
In summary, GP is an attractive surrogate model for BO, which allows for the quantification of uncertainty in predictions. It is a nonparametric model and its number of parameters only depends on the input points. With a proper kernel function, GP is able to take advantage of the data structure. However, GP also has some drawbacks. For example, it is conceptually difficult to understand along with the theory of BO. Furthermore, its poor scalability with high dimensions or large number of data points is another key problem~\citep{feurer19}. Most kernels have hyper-parameters that determine the prior of a Gaussian process. Moreover, the choice of prior highly influences the performance for a given amount of data. GP itself is a computational resource-consuming process with the $O(N^3)$ computation cost. Computing the kernel matrix costs $O(DN^3)$ with $O(N^2)$ memory~\citep{murray16}.\\
\indent
Acquisition function is a function of data point x, designed to choose the next sampling point. Acquisition functions must be carefully selected to trade off exploration over the search space and exploitation in current areas. The most common choice is the expected improvement (EI) because of its good performance and ease of use~\citep{frazier18}. Its mathematical format is as follows:\\
\[
    E[I(\lambda)]\;=\;E[max(f_{min}-y,0)]
\]
\indent
The improvement function is defined as $I(x,v,\theta)=(v-\tau)I$. \emph{I} is larger than zero when an improvement exists; \emph{v} is a normally distributed random variable; and \emph{$\tau$} s the target. With the function of improvement \emph{I}, the analytical expression of EI is as follows:\\
\[
    E[I(\lambda)]\;=\;E[I(x,v,\theta)]\;=\;\Big(\mu_n(x)-\tau\Big)\Phi\Big(\frac{\mu_n(x)-\tau}{\theta_n(x)}\Big)+\theta_n(x)\phi\Big(\frac{\mu_n(x)-\tau}{\theta_n(x)}\Big)
\]
where $\sigma_n(x)>0$, $\phi(\cdot)$ and $\Phi(\cdot)$ are the standard normal probability density function and cumulative standard normal distribution function; and $\mu_n(x)$ is the best observed value thus far~\citep{jones98}. In addition to improvement-based policies, knowledge gradient function~\citep{frazier09}, entropy search function~\citep{hennig12}, and predictive entropy search function~\citep{hernandez14} have recently been developed to enhance performance over different applications. A portfolio with acquisition functions was also proposed to cope with different strategies. It could be designed based on the hedge algorithm, using which the acquisition function was chosen with past performance~\citep{hoffman11}. An entropy search portfolio~\citep{shahriari14}  selects the acquisition function using the information gained toward optimization.\\

\indent
BO is efficient because it selects the next hyper-parameter set in a guided manner. With BO, users will make fewer calls to the objective function. For a machine learning model, BO is more efficient at finding the optimum hyper-parameter compared with grid or random search. However, there are some downsides of the vanilla Bayesian method. The most prominent problem is that BO is a sequential process in which a trial is proposed using experience from previous trials. Some solutions have been raised to solve the parallelism problem. A method was proposed based on the batch input of current results~\citep{ginsbourger11}. As GP+EI is the most popular combination for Bayesian optimization, improvements based on EI~\citep{ginsbourger10,snoek12} and GP-UCB~\citep{hutter12, desautels14} Parallelization on other acquisition functions (e.g. information-based function) and the combination of BO with other algorithms~\citep{falkner18} are also feasible solutions.\\
\indent
Another drawback of BO is the greater consumption of computational resources compared with grid search and random search. This especially draw our attention when it is applied to deep learning networks. Memory consumption, training time, and power consumption are problems for training a neural network. Therefore, the search algorithm not only needs to be fast but also resource-efficient. A straightforward method for solving this problem is to set a boundary and determine whether a certain trial is still worthy of training. This “boundary” could be the loss function or average accuracy in the neural network. Domhan roposed a predictive termination method that can be combined with the Bayesian method. This idea can be combined with search algorithms other than BO~\citep{domhan15}.Freeze-thaw Bayesian optimization~\citep{swersky14} utilizes a forecast curve to determine whether to freeze a current trial or thaw a previous one. Currently, BO has been applied for hyper-parameter tuning in deep learning networks, which has led to achievements in both image classification~\citep{snoek15} and natural language processing~\citep{melis17}.\\
\indent
Bayesian methods were originally not applicable for integers, categorical values, or a conditional search space. However, in reality, hyper-parameters for deep learning could be in any data type. For vanilla Bayesian, kernel function and optimization process must be adapted to involve integers (e.g., batch size) and categorical data~\citep{hutter09,garrido17}. Conditional variables may only influence the results with some other variable taking certain values. This is common in deep learning networks, for example, hyper-parameters related to certain optimizers. Models with a tree-based structure have been designed to solve this problem. In addition to GP, random forests~\citep{hutter11} are another popular type of surrogate model with a structure similar to a decision tree. Tree-structured Parzen Estimators~\citep{bergstra11,bergstra2013making} are discussed in depth in the following subsection.\\

\subsubsection{Tree Parzen Estimators}
Tree Parzen Estimators (TPEs), models with a graphic structure that deal with the conditional search space, are an alternative choice for surrogate model. In the regular Bayes rule, the surrogate model is represented as $p(y|\lambda)=\frac{p(\lambda|y)p(y)}{p(\lambda)}$ of observation y given configuration. Furthermore, $p(\lambda|y)$ is the probability of hyper-parameters given the score on the objective function. TPE convert $p(\lambda|y)$ into a tree-structured expression:\\
\[
p(\lambda|y)\;=\;\left\{
        \begin{array}{rcl}
        l(\lambda), & y\;<\;0\\
        g(\lambda), & x\;\geq\;0\\
        \end{array}
        \right.
\]
where \emph{$\alpha$} is the threshold dividing observations into “good” and “bad”. The percentage is usually set to 15\%, and therefore the hyper-parameters have two different distributions, $l(\lambda)$ and $g(\lambda)$. Instead of directly drawing a value from $l(\lambda)$, TPE uses the ratio of $l(x)/g(x)$ to maximize the expected improvement acquisition function:\\
\[
    EI_{\alpha}(\lambda)\;=\;\frac{\gamma \alpha l(\lambda)-l(\lambda)\int_{-\infty}^{\alpha}p(y)dy}{\gamma l(\lambda) + (1-\gamma)g(\gamma)} \varpropto \Big(\gamma+\frac{g(\lambda)}{l(\lambda)}(1-\gamma) \Big)^{-1}
\]
In summary, the workflow of TPE is as follows: draw a sample from $l(x)$, evaluate $l(x)/g(x)$ with EI, and obtain the optimum value of $l(x)/g(x)$ with the greatest EI. TPE outperforms the Bayesian method in dealing with conditional variables because of its tree structure~\citep{strobl08}.\\
\indent
HyperOpt~\citep{komer14} realizes asynchronous parallelization and uses TPE to facilitate the HPO. BOHB combined TPE with HyperBand and successfully solved the problems of parallelization and conditional space. TPE is even more popular than the original Bayesian method when applied to deep learning networks because it is applicable for more data types. However, GPs still performance better when hyper-parameters have strong interaction because TPE do not model the interactions~\citep{bissuel18}.\\

\subsection{Optimization with an Early-stopping Policy}
HPO is often a time-consuming process that is computationally costly. In a realistic scenario, it is necessary to design the HPO process with limited available resources. When experts tune hyper-parameters by hand, they are occasionally able to use their experience of parameters to narrow down the search space and evaluate the model during training, and determine whether to halt the training or continue it. This is especially common for neural networks because it takes a long time to train a model with each hyper-parameter set. An early stopping strategy is a series of methods that mimics the behavior of AI experts to maximize the computational-resource budget for promising hyper-parameter sets.\\
\indent
In addition to efficient search algorithms, a strategy to evaluate trials and determine whether to halt them early is another active area of research. The following early-stopping algorithms can be applied in combination with a search algorithm or individually for both searching and halting. The early-stopping algorithms for HPO are similar to that for neural network training, but for neural network training, early stopping is applied to avoid overfitting. This allows users to terminate a trial before finishing the whole training, thereby freeing computational resources for trials with a promising hyper-parameter set.\\

\subsubsection{median stopping}
Median stopping is the most straightforward early termination policy, adapted by structured HPO frameworks such as Google Vizier~\citep{golovin17}, Tune~\citep{liaw18}, and NNI~\citep{msr18}. It is model-free and applicable to a wide range of performance curves. Median stopping makes a decision based on the average of primary metrics (e.g., accuracy or loss) reported by previous runs. A trial \emph{X} is stopped at step \emph{S} if the best objective value by step \emph{S} is strictly worse than the median value of the running average of all completed trials’ objective values reported at step \emph{S}~\citep{golovin17}. The median stopping strategy is not a model, and thus no hyper-parameters need to be determined by users.\\

\subsubsection{curve fitting}
\begin{figure}
    \centering
    \includegraphics[width=14.5cm,height=8cm]{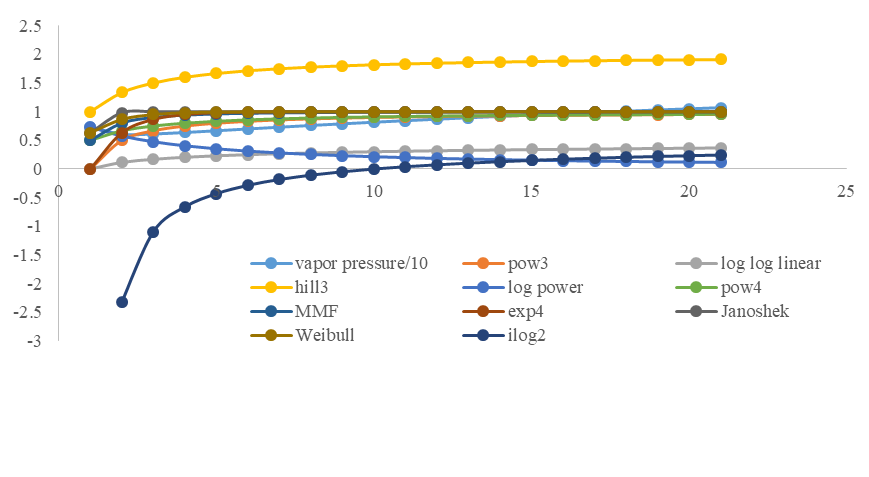}
    \caption{Functions used for curve-fitting, with related hyper-parameters $a=1$, $b=1$, $\alpha=1$, $\beta=0$, $\kappa=1$, and $\delta=1$ or 2.}
    \label{Figure 13}
\end{figure}

Curve fitting is an LPA (learning, predicting, assessing) algorithm~\citep{kohavi95,provost99}. It is applicable in the combinations of search algorithms mentioned in Section 2, supported by Google Vizier and NNI. This early stopping rule makes a prediction of the final objective value (e.g., accuracy or loss) with a performance curve regressed from a set of completed or partially completed trials. A trial \emph{X} will be halted at step \emph{S} if the prediction of the final objective value is sufficiently worse than the tolerant value of the optimal in the trial history.\\
\indent
This algorithm has been commonly supported by Bayesian parametric regression~\citep{swersky14,golovin17}. A method is proposed~\citep{domhan15} that used a weighted probability learning curve with a combination of 11 different increasing saturated curves (Figure \ref{Figure 13}). The combination model can be expressed as:\\
\[
    f_{comb}(t|\xi)\;=\;\sum_{k=1}^K w_kf_k(t|\theta_k)
\]
with new combined parameter vectors\\
\[
    \xi\;=\;(w_1,\ldots,w_K,\theta_1,\ldots,\theta_K,\sigma^2)
\]
where \emph{$w_K$} is the weight for each model \emph{k}, \emph{$\theta_K$} is the individual model parameter, and \emph{$\sigma^2$} is the noise variance.\\
In general, there are three steps to determine whether to stop:
\begin{itemize}
\item[-] Learning: The parameters of the curve are learned from current completed trials: first, fit each curve with the least squares method and obtain parameter \emph{$\theta_K$}. Then, filter the curve and remove outliers. Finally, adjust the weight \emph{$w_K$} for each curve using the Markov Chain Monte Carlo sampling method.
\item[-] Predicting: Calculate the final objective value at a certain step, with the curve obtained from the first step. 
\item[-] Assessing: If the predictive result does not converge, the model will continue to train, obtain further information, and make a judgment again. If the predictive result is superior within a threshold of historical best values, the result will be retained and training stopped; otherwise, the result will be discarded and the training stopped.
\end{itemize}
\indent
Compared with median stopping, the curve fitting method is a model with parameters. Building the model is also a training process. When combined with search algorithms, the predictive termination accelerates the optimization process and then finds a state-of-the-art network. As mentioned previously, Freeze-Thaw BO can be viewed as a combination of BO and curve fitting.\\

\subsubsection{Successive Halving and HyperBand}
This and the next subsection discusses several bandit-based algorithms with strong performance in optimizing deep learning hyper-parameters. HPO in deep neural networks is more likely to be a tradeoff between accuracy and computational resources because training a DNN is a very costly process. Successive halving (SHA) and HyperBand outperform traditional searching algorithms without early stopping in saving resources for HPO, with random search as a sampling method and a bandit-based early stopping policy.\\
\indent
SHA~\citep{jamieson16} is based on multi-bandit problems~\citep{karnin13}. SHA converts HPO into a nonstochastic best-arm identification to allocate more computational resources to more promising hyper-parameter sets. Bandit processes are a special type of Markov Decision Process with several possible choices~\citep{kaufmann17}, and they have a long history of stochastic setting. K-armed bandit is an efficient tool for dealing with sequential decisions over time with uncertainty~\citep{slivkins19} and stopping rules based on the Gittins index theorem~\citep{gittins74}.The SHA model was built upon the multi-armed bandit formulation~\citep{agarwal12,sparks15}.\\
\begin{figure}
    \centering
    \includegraphics[width=12.3cm,height=6.2cm]{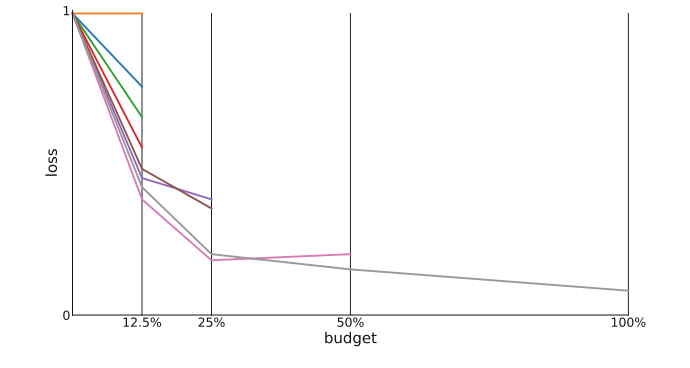}
    \caption{Budget allocation with successive halving~\citep{bissuel18}}
    \label{Figure 14}
\end{figure}
\indent
The SHA algorithm can be described as follows: users need to set an initial finite budget \emph{B} and the number of trials \emph{n}, uniformly query all hyper-parameter sets for a portion of initial budget, evaluate the performance of all trials, drop the worse performing half, double the budget for the remaining half, and repeat the whole pipeline until only one trial remains (Figure \ref{Figure 14}). For training a deep learning network, the budget could be the number of iterations or the total training time.\\
\indent
Compared with BO, SHA is theoretically easier to understand and more computationally efficient. Instead of evaluating models when they are fully trained to convergence, SHA evaluates the intermediate results to determine whether to terminate it. The major drawback of SHA is the allocation of resources. Given a certain budget, a tradeoff will occur between total budget \emph{B} and the number of trials \emph{n}(the “\emph{n} vs. \emph{B/n}" problem), which are both decided by users in advance. With an overlarge \emph{n}, each trial will be given a small budget that may result in premature termination, while an insufficient \emph{n} would not be able to provide enough optional choices. Furthermore, too large a budget may lead to a waste of choice, and too small a value may not promise an optimum.\\

\indent
HyperBand (HB)~\citep{li17} s an extension of SHA, mainly designed to solve the “\emph{n} vs. \emph{B/n}" problem by considering several possible \emph{n} values with a fixed \emph{B}. Similar to SHA, it also formulates the HPO into a pure-exploration, nonstochastic, infinite-armed bandit problem. HB has two improvements because it adds an out loop over the routine of SHA. The first modification is on the allocation of resources \emph{B}. n out loop iterates over different combinations of \emph{n} and \emph{B}, and thus different tradeoffs will occur between \emph{n} and \emph{B/n}. Users must set a maximum amount of resource \emph{R} that can be allocated to a single trial to determine how many different combinations need to be considered. The other modification is the proportion of the trial to be discarded \emph{$\eta$}. Instead of 50\%, the worst faction $\frac{\eta-1}{\eta}$ will be dropped and the budget for remaining trials multiplied by \emph{$\eta$}. This is another hyper-parameter decided by users. The default value of \emph{$\eta$} is 2 in SHA and 3 or 4 in HB. The suggested value and explanation of \emph{R} and \emph{$\eta$} were explicitly presented in the publication of HB~\citep{li17}. \emph{R} has a limitation based on computation or memory resources. Because the number of combinations \emph{S} is a function of \emph{R} ($S_{max}=\lfloor{log_n(R)}\rfloor$), a smaller \emph{R} usually indicates faster evaluation while a larger \emph{R} gives a better guarantee of finding the optimal configuration. The value of \emph{$\eta$} can be determined by users within a specific limit. A larger \emph{$\eta$} implies a more aggressive rate for elimination and the user will receive the results faster.\\
\indent
HB is easy to deploy in parallel because all trials are randomly sampled and run independently. It accelerates random search through the early stopping inherited from SHA and a more reasonable adaptive allocation of resources. Compared with random search and BO, HB exhibits superior accuracy with less resources, especially in the case of stochastic gradient descent for deep neural networks. Currently, HB is involved in Tune and NNI as part of trial scheduler for the HPO framework.\\

\subsubsection{Asynchronous Successive Halving and Bayesian Optimization–HyperBand}
The paper on HB~\citep{li17} concludes by discussing potential improvements, including distributed implements, adjustment of the convergence rate, and non-random sampling. Methods discussed in this subsection can be viewed as extensions of HB. Asynchronous SHA (ASHA)~\citep{li18} proposes an enhanced distribution scheme for parallelizing HB and avoids straggler issues during elimination. As mentioned previously, Bayesian Optimization–HyperBand (BOHB)~\citep{falkner18} is a combination of BO and HB, and introduces a guided sampling method instead of simple random search (Table 2).\\
\indent
ASHA modifies the original HB by improving the efficiency of asynchronous parallelization. It has similar performance to HB in sequential setting or small-scale distribution. For large-scale parallelization, ASHA exceeds most state-of-the-art HPO methods. The computational resources are automatically allocated by ASHA, and thus there are no extra hyper-parameters for users compared with HB. The currently published ASHA is still an incomplete version, and it is incorporated into Tune as a trial scheduler.\\
\begin{table}
    \centering
    \begin{tabular}{lllll}
     \toprule
     &  SHA  & HB & BOHB & ASHA \\
     \midrule
     Discard ratio & half & adjusted by user (1/3)  & adjusted by user & adjusted by user\\
     B vs n & one choice & several combinations & several combinations & several combinations \\
     Sampling & random &  random & TPE  & random\\
     Parallelization & synchronous & synchronous &  synchronous & asynchronous \\
     Scalability & small scale & small scale & small scale & large scale \\
     Time & 2015 & 2018 & 2018 & 2018 \\
     \bottomrule
    \end{tabular}
    \caption{Comparison between bandit-based algorithms}
    \label{Table 2}
\end{table}
\indent
HB is successful in allocating trials and resources but limited by random sampling, especially in the final few steps for HPO. Random search is effective in the early stage of sampling, but ineffective when approaching the optimum configuration. Several approaches combining BO and HB have been proposed in recent years~\citep{bertrand17,falkner18,wang18}, to solve the native problem with HB. Among them, BOHB is the most widely applied as it is the only open-source one. The choice of budgets and schedule of trials are inherited from the original HB. The BO part closely resembles the TPE method~\citep{bergstra11} with a single multidimensional Kernel Density Estimation (KDE). Compared with the hierarchy of one-dimensional KDEs used in the original TPE, the improved version can better handle interaction effects. BOHB achieves quick convergence to the best hyper-parameter set with Bayesian methods, and the feasibility of parallelization with HB.\\
\begin{figure}
    \centering
    \includegraphics[width=12cm,height=6cm]{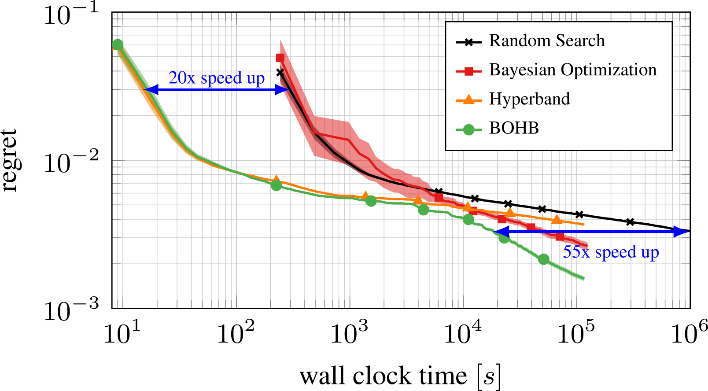}
    \caption{Comparison between BOHB and HyperBand with different budgets~\citep{bissuel18}}
    \label{Figure 15}
\end{figure}
\indent
Advantages inherited from HB include budget control, the ability to check performance at any time, computational efficiency in the early stage of tuning, and – most crucially – scalability. The involvement of BO ensures strong final performance, robustness, and computational efficiency in the later stage of tuning. The empirical performance of BOHB demonstrates state-of-the-art results over a wide range of HPO tasks, including Bayesian neural networks, reinforcement learning agents, and convolutional networks. From the results shown in Figure \ref{Figure 15}, BOHB can be noticed to mainly benefit from the strategy of HB if given a limited budget, and both exhibit 20-times acceleration over random search. However, with more sufficient budgets, BOHB outperforms HB because of the guided sampling strategy with the Bayesian method, exhibiting 55-times acceleration over random search. The original open-source version is HpBandSter on GitHub. In general, BOHB is a robust and computationally effective HPO method that is easy to apply in multiple units.\\
\indent
The major shortcoming of BOHB lies in the setting of budgets. Automatic adaption of budgets may solve the problem especially for users with less experience. The application of BOHB requires users to have some experience. In some extreme cases, either BO or the trail schedule borrowed from HB could be a drawback for efficient searching. Compared with random search, Bayesian methods may take longer to escape the local optimum~\citep{biedenkapp18}. Another downside is inherited from bandit-based methods: users must define a proper budget that matches the required accuracy, but this is not an easy task because it needs sufficient experience of these algorithms. If a smaller budget is too noisy to make any decision on the model performance, adjustments to the budget plan are necessary, which is a time-consuming process. For this reason, BOHB could be several times slower than vanilla TPE, and the rate is influenced by the “\emph{n} vs. \emph{B/n}" problem. Setting a small budget is risky if one is not fully familiar with the training process or tuning algorithms. When the downside of bandit-based methods encounters the trap in the local optimum, BOHB may be less efficient compared with other optimization methods; however, this only happens in some extreme cases.\\

\subsubsection{Population-Based Training}
\begin{figure}
    \centering
    \includegraphics[width=14.6cm,height=6.4cm]{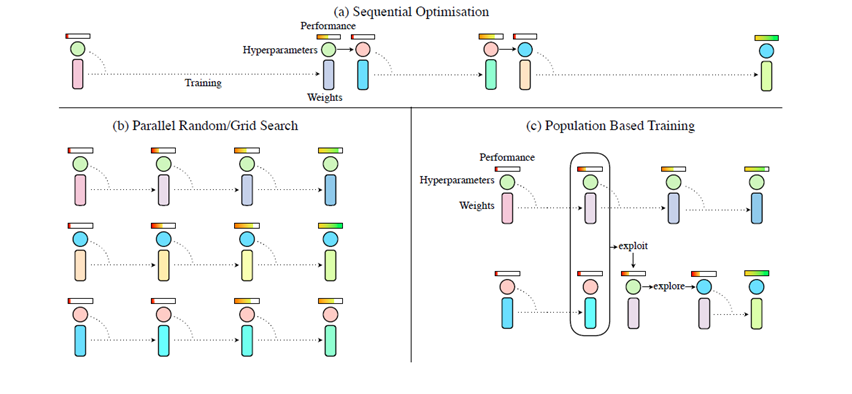}
    \caption{PBT as a combination of sequential and parallel optimization ~\citep{jaderberg17}}
    \label{Figure 16}
\end{figure}

Population-based methods are essentially series of random search methods based on genetic algorithms~\citep{shiffman12}, such as evolutionary algorithms~\citep{simon13,orive14}, particle swarm optimization~\citep{eberhart98,lorenzo17}, and covariance matrix adaption evolutionary strategy~\citep{hansen16}. The most important parts of genetic algorithms are initialization (random creation of a population), selection (evaluation of the current population and selection of parents), and reproduction (creation of the next generation). One of the most widely used population-based methods is population-based training (PBT)~\citep{jaderberg17,li19} proposed by DeepMind. PBT is a conceptually simple but computationally effective method. It is a unique method in two aspects: it allows for adaptive hyper-parameters during training and it combines parallel search and sequential optimization. These features make PBT especially suitable for HPO in deep learning networks.\\
\indent
The process of PBT can be simply described as being similar to generic algorithms. First, a population of trials with different hyper-parameter settings is initialized (Figure \ref{Figure 16}). The size of the population is determined by the user (20–40 has been suggested in a publication) and all neural networks are trained in parallel. Then, for every \emph{m} iterations, all models are evaluated with a certain metric, which could be loss or median accuracy. The parallelization is inherited from traditional genetic algorithms. Finally, each trial uses the information from the rest of the population to update the hyper-parameters through exploitation and exploration; their balance is similar to the concepts of hand tuning and BO. Exploitation of the best configuration is defined by replacing the current weights with weights that have the best performance. This process is similar to inheritance and crossover in genetic algorithms. Exploration is ensured by the random perturbation of hyper-parameters with a noise defined by the user. This process is similar to mutation in generic algorithms, and the noise can also be described as the mutation rate. Exploitation and exploration are performed periodically, and training will continue with the described steps.\\
\indent
PBT provides a way to involve HPO in regular training, using a warm start instead of waiting for convergence. This is meaningful for large models, such as the training for a generative adversarial network (GAN) ~\citep{goodfellow14} and a transformer network~\citep{vaswani2017attention}. It takes a long time to train these large models to convergence. In this case, PBT will be far more effective than the abovementioned HPO methods, and it is the only algorithm using transformer as a show case to prove its effectiveness. For the optimization of adaptive hyper-parameters (e.g., LR), it is unnecessary for users to decide whether to use linear decay or exponential decay because PBT will change the values of all hyper-parameters periodically.\\
\indent
There are still some drawbacks to be explored in the future. For example, this method is not theoretically proven to obtain the optimal hyper-parameter set. It is convincing that PBT provides the best configuration with current computational resources, but whether it is the optimum configuration is still open to discussion. Another problem is the comparatively easy strategy for exploitation and exploration. PBT is still not extendable to advanced evolution and mutation decisions. In addition, the definition of hyper-parameters and the changes made to the computation graph are complicated.\\

\indent
Algorithms can be viewed as theories supporting HPO. For practical application, users always need to decide which hyper-parameters take into consideration, the way to apply the search algorithms and the process to train deep learning models. In this case, a tool is necessary to conduct these configurations.\\

\section{Toolkits for Hyper-parameter Optimization}
Hyper-parameter training is a time-consuming process, especially for deep learning networks. It may take decades of GPU days to finish the whole process because training a single neural network to convergence will take a whole day. A toolkit for HPO builds a bridge between network training and hyper-parameter tuning. The combination of tuning algorithm, scheduler for trials, and compatibility with major deep learning toolkits expedites the optimization process and lowers the threshold for research \& development.\\
\indent
This section provides a general comparison of contemporary open source toolkits and services, and discusses the most integrated ones in detail. The frameworks are assessed by their ability to handle resources, support state-of-the-art optimization algorithms, and schedule for trials.\\

\subsection{Overview}
Generally, there are two types of toolkits for HPO: open source tools and services that rely on cloud computing resources, and each of which are described in this subsection.\\
\indent
Multiple open-source libraries have been created to cater to the demand for different tasks in automatic model design, including model construction, feature engineering, and HPO. Some of them were designed for the application of the proposed algorithm. For example, HyperOpt is specifically designed for asynchronous BO (TPE) based on Gaussian processes and regression trees. HpBandSter features in the implementation of BOHB, although it also provides random search and HB. Xcessive outperforms other libraries with an interactive graphical user interface (GUI) that is beginner-friendly. Users are able to manage the training, optimization, and evaluation of models through the GUI. However, Xcessive only supports automated hyper-parameter search through Bayesian methods, and it is also inconvenient for training deep learning networks. Scikit-Optimize is a comprehensive library cited by many later frameworks. It contains Bayesian methods, random search, random forest, and some other reliable optimization algorithms. All aforementioned methods have downsides in common: their implementation on search algorithms, their support for deep learning training frameworks, and their application schedule for trials. These inhibit efficient search for hyper-parameters for deep neural networks.\\
\indent
In addition, Google Cloud and Amazon Web Services (AWS) provide model selection and HPO. These services are supported by massive computational resources, applicable for thousands of parallel training processes and evaluations. However, they are a closed-source infrastructure and users must pay for the service.\\
\newcommand{\tabincell}[2]{\begin{tabular}{@{}#1@{}}#2\end{tabular}}
\begin{table}
    \centering
    \begin{tabular}{cll}
     \hline
     & \textbf{Vizier \& Sagemaker} & \textbf{NNI \& Tune}\\
     \hline
     {Ease to use} & \tabincell{l}{Almost no configuration by users\\- Friendly to starters with simple \\workflow\\- Straightforward GUI for both task \\management and result visualization
    \\- Error correcting capability} & \tabincell{l}{- Need minimal configuration \\by users \\ - Preliminary knowledge on model\\ training\\ - Visualization with Tensorboard}\\ 
     \hline
     Scalability & \tabincell{l}{- Very high with the support of \\cloud service} & \tabincell{l}{- Usually deployed on several units}\\
     \hline
     {State-of-the-art} & \tabincell{l}{- Support classical search algorithms\\- Some early stopping methods\\- Transfer learning } & \tabincell{l}{- Support almost all SOTA algorithms\\- Not support transfer or meta \\learning }\\
     \hline
     {Availability} & {- Need to pay for the convenience} & {- Open sourced and free}\\
     \hline
     {Flexibility} & \tabincell{l}{- Extendable search algorithm\\ - Closed sourced in infrastructure} & \tabincell{l}{- Able to modify everything\\ - High flexibility for experienced users } \\
     \hline
    \end{tabular}
    \caption{Comparison between closed-source services and open-source toolkits}
    \label{Table 3}
\end{table}
\indent
The tools for HPO are designed to satisfy the following considerations: ease of use, state of the art, availability, scalability, and flexibility~\citep{golovin17}. Table \ref{Table 3} displays a general comparison of the open source tools and services in regard to the abovementioned considerations. More details are discussed in the following subsections.\\

\subsection{Google Vizier}
Google Vizier~\citep{golovin17} is a scalable service instead of a library for black-box optimization based on Google’s Cloud Machine Learning subsystem. The most prominent advantage is its ease of use. With the service, users only need to choose an involved search algorithm and submit a configuration file, and then they will be provided with a suggested hyper-parameter set. All other setup tasks have been conducted by Google Vizier, such as the deployment of the system, management of computational resources and memory, and scheduling of trials. The simple client workflow and minimal configuration make it especially friendly for starters (Figure 17), while for experienced users it also allows them to customize their algorithm with the algorithm playground. Although Google Vizier is closed-source in infrastructure, it is easy to change or design new algorithms for HPO.\\
\begin{figure}
    \centering
    \includegraphics[width=10cm,height=5.5cm]{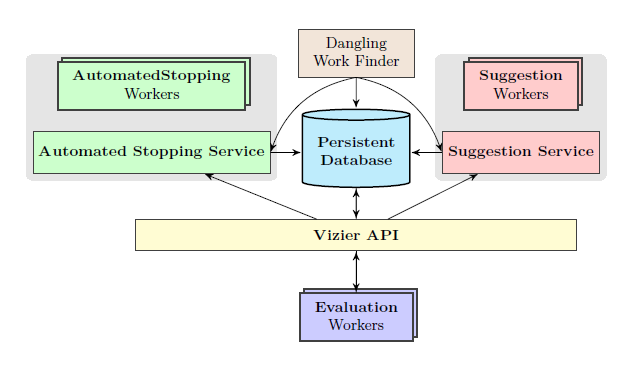}
    \caption{Architecture of Google Vizier~\citep{golovin17}}
    \label{Figure 17}
\end{figure}
\indent
Another advantage over open-source libraries is the very high scalability. Google Vizier is scalable to millions of experiments and billions of trials with different hyper-parameter sets. All jobs are partitioned across Google data centers and processed by Google’s load-balancing infrastructure. The whole system – with a strong error-correcting ability – will pause a trial if it fails and prevent it from affecting the stability of the whole job.\\
\indent
A third benefit is the implementation of transfer learning, which helps users to accelerate their tuning processes but rarely implemented in an open-source library. This strategy, built with GP regression, is based on previous jobs submitted by millions of users. Instead of trained from scratch, the model submitted by user will be fine-tuned with provided dataset. Transfer learning is often more valuable for deep learning networks because it is expensive to train.\\
\indent
In addition, Google Vizier provides a benchmarking suite to measure the efficiency of algorithms with performance-over-time metrics.\\
\indent
Additional reason to choose Vizier is the GUI, which is applicable for both monitoring the status and changing the setting for an experiment. There is almost no technical threshold for common users because they are not even required command lines to create an experiment or request a new suggestion.\\
\indent
Vizier does not outperform in search algorithms and early stopping strategies. Similar to most open-source libraries, Google Vizier supports float, integer, discrete, and categorical parameters at a logarithmic scale. Grid and random search are involved, while batched Gaussian process bandit optimization~\citep{desautels14} is the default mode for most scenarios. Mat\'ern kernel with an automatic relevance determination~\citep{rasmussen03} and expected improvement acquisition function~\citep{mockus78} are applied. The early stopping algorithms are Curve Stopping Rule (refer to 4.2.2) and Median Stopping Rule (refer to 4.2.1). In a recent study by DeepMind~\citep{li19},  Vizier was implemented with a PBT framework. This work indicated the feasibility of implementing new optimization algorithms in Vizier, and the support of Vizier’s infrastructure was helpful in accelerating the tuning process. The suggested search algorithm and default values for hyper-parameters makes it friendlier for users with less experience, whereas with most open-source frameworks users must make decisions based upon their own experience.\\
\indent
In general, Vizier is a forerunner of a comprehensive HPO service. The key design goal is the significant reduction of setup efforts. As a beginner in hyper-parameter tuning, one can start and monitor an experiment with the dashboard, suggested with an efficient search algorithm and early stopping strategy, use the allocation of Google Cloud, and finally check the results on the UI. Very little effort is required if one is willing to use the default settings provided by Vizier. Advanced users are able to replace the algorithm with their own design, but they may still perceive a lack of flexibility. All in all, if deploying a framework and setting up an experiment generate no problems, one may not deem paying for these features to be worthwhile.\\
\indent
Advisor is an open-source version of Google Vizier~\citep{advisor18} that attempts to realize all of Vizier’s features without the support of Google Cloud. Grid search, random search, and BO are also supported in Advisor, as are early stop algorithms. Advisor even realizes an interactive GUI, allowing users to easily start and stop an experiment. However, the open-source version will never reproduce the extremely high scalability of Vizier because it is limited by the availability of computational resources.\\

\subsection{Automatic model tuning in Amazon SageMaker}
Automatic model tuning is a module of Amazon SageMaker, a machine learning environment for simplifying the process of model building and deployment. As a service, it is similar to Google Vizier with the support of Amazon Web Services (AWS). The most attractive feature is also its ease-to-use: only a model and related training data are required for a HPO task. It supports optimization for complex models and datasets with parallelism on a large scale. The involvement of Jupyter simplifies the configuration for optimization and visualization of results. The HPO process and status of trials can be viewed with the Amazon Sagemaker console, which supports early stop and warm start. Search algorithms are limited to random search and BO, but users are provided with enough freedom to add their own algorithms. Amazon SageMaker’s automatic model tuning shares most of the advantages and limits of Google Vizier because both of them are cloud-based services. The major difference is that Amazon SageMaker, as a fully managed service, integrates every step of training a neutral network; HPO is just a module of this service. It is not necessary to switch to different tools to build, train, and tune a model.\\

\subsection{Neural Network Intelligence}
\begin{figure}
    \centering
    \includegraphics[width=11cm,height=5.36cm]{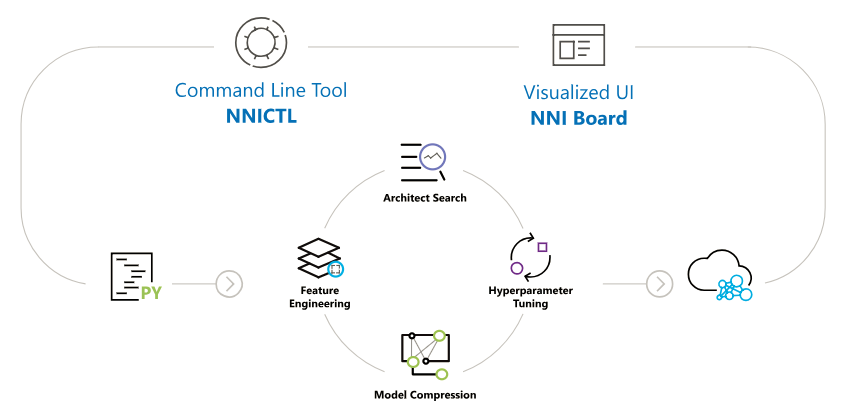}
    \caption{Structure of Neural Network Intelligence~\citep{msr18}}
    \label{Figure 18}
\end{figure}

\begin{figure}
    \centering
    \includegraphics[width=14cm,height=7.36cm]{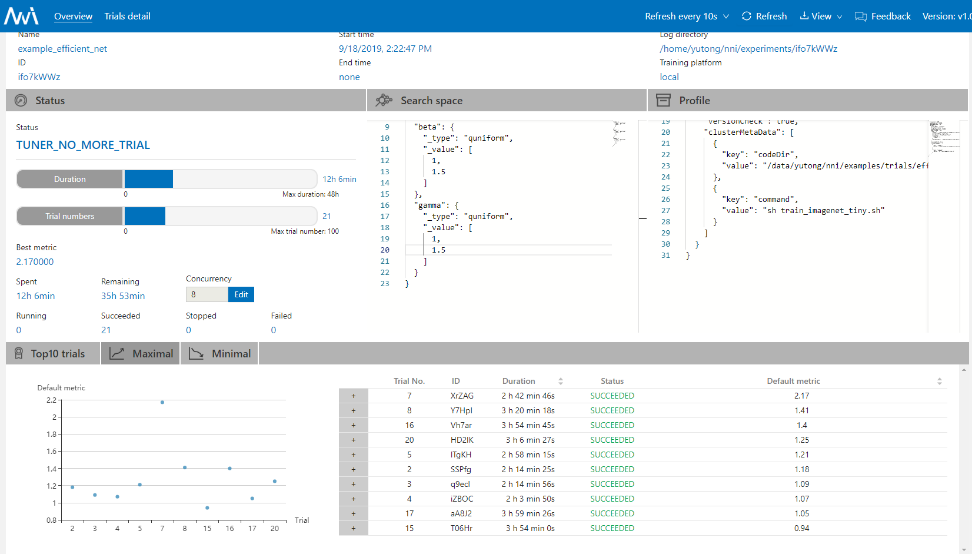}
    \caption{Example of Web UI for NNI}
    \label{Figure 19}
\end{figure}

Neural Network Intelligence (NNI) is an open-source toolkit for both automated machine learning (AutoML) and HPO released by Microsoft. Compared with the abovementioned closed-source services, NNI provides not only a framework to train a model and tune hyper-parameters but also more freedom for customization (Figure \ref{Figure 18}). This toolkit can be deployed in different environments such as local machines, remote servers, and Dockers. The support from Microsoft provides promising maintenance and update potential. In general, it is more suitable for researchers with some knowledge of machine learning if they are willing to (a) perform the deployment job by themselves; (b) test more algorithms or networks in a different environment; and (c) involve HPO or AutoML in the NNI platform.\\
\indent
This section discusses the main advantages and drawbacks of NNI. NNI implements more search algorithms than Google Vizier and SageMaker, and provides an extensible interface to write a new algorithm. In NNI, all algorithms related to HPO are generally called “Tuners” or “Advisors”, and “Assessors” contain two early stopping algorithms. Actually, there are two groups of algorithms. Tuner is similar to search algorithms, as mentioned in Section 4.1, including grid search, random search, TPE, and their variants. The other group refers to the schedulers mentioned in Section 4.2, including HB and BOHB. Almost all state-of-the-art algorithms (except PBT) are built into NNI. If a user wishes to explore the efficiency of different algorithms, NNI is a convenient choice. Tuners can be applied in combination with Assessors. Advisors already have their own early stopping strategy, and thus Assessors are not applicable. All configurations for Tuners, Assessors, and Advisors are set in the \emph{config.yml} file.\\
\indent
In addition, NNI is designed with high extensibility. As well as the built-in Tuners and Assessors, researchers can test new self-designed algorithms. They only need to inherit the base Tuner/Assessor, implement related functions, and configure the new algorithm into the \emph{config.yml} file.\\
\indent
NNI is able to be deployed in all kinds of infrastructure, including local machines, remote servers, and Kubernetes-based Dockers. It is extensible to implement more machine learning frameworks, both by users and developers of NNI. In addition, NNI is compatible with most popular deep learning frameworks and libraries, including PyTorch, Keras, TensorFlow, MXNet, Scikit-learn, and XGBoost. There are also extensible points for new platforms in NNI.\\
\indent
When an HPO experiment is initialized, its process and intermediate results can be monitored through a visualization interface named Web UI (Figure \ref{Figure 19}). Web UI displays almost all information in need of AutoML and HPO, including the status of the experiment and each trial, search space and configuration information, log, intermediate results, plots between metrics and hyper-parameters, plots showing relations between hyper-parameters, and comparisons between different trials. TensorBoard and TensorBoardX are also compatible with NNI. Compared with Web UI, plots are created by command lines in TensorBoard, which is more flexible but also requires more knowledge.\\
\indent
Documents and examples are helpful for starters. Examples for all search algorithms, frameworks, and hardware platforms are prepared to lower the technical threshold. NNI’s developers are still working on the implementation of more algorithms and platforms, which guarantees that it will be updated for SOTA development in AutoML.\\

\indent
However, NNI still has some downsides. Most AutoML/HPO tools are designed for a specific algorithm (e.g., TPE) or certain tasks (e.g., image classification). The advantage is that users only need to provide labeled data, and then the model training will be performed by tools. The complexity of the model is wrapped up using HPO tools, and therefore knowledge for model training is not a must. Compared with most existing tools and services, NNI is universal for all training and optimization tasks, with any machine learning framework on any platform. This is definitely an advantage, but could be a drawback for beginners because model training experience is a prerequisite.\\
\indent
Furthermore, the function of Web UI could be extended. Compared with Google Vizier, NNI only displays the status and results of experiments with no interactive function. Neither the hyper-parameter setting, choice of algorithm, or configuration of resources can be adjusted from Web UI.\\
\indent
A third problem lies in debug information. Specifically, the debugging process could be more straightforward. Users may need more details in the log file, especially when the NNI manager or dispatcher fails.\\
\indent
NNI supports almost all cutting-edge search algorithms except for population-based algorithms. PBT is quite a unique algorithm in its combination of the process of hyper-parameter tuning and model training. It is especially efficient for models on a very large scale.\\

\subsection{Ray.Tune: An open-source toolkit for hyper-parameter search}
Ray.Tune is a library of Ray developed by Berkeley RISELab. As a distributed framework for model training, Ray guarantees the efficient allocation of computational resources. Most searching methods implemented in Tune can be realized in parallel. As an open-source framework, Tune shares similarities with NNI in many applications, but there are also some major differences (Table \ref{Table 4}), which are listed as follows:\\
\begin{itemize}
\item[-]NNI is a toolkit for both AutoML and HPO, whereas Tune is a library as a part of Ray only supporting hyper-parameter tuning.
\item[-]Currently, Tune is not applicable for Kubernetes-based Dockers.
\item[-]Tune support PBT, which is a great advantage over NNI.
\item[-]Tune implements most search algorithms by importing existing libraries (e.g., BayesOpt and HyperOpt), whereas the developers of NNI rewrite most of them. Therefore, users may need to install the relevant package when using a certain algorithm.
\item[-]To start an experiment, users must define a trainable class and all configurations are set in a tune.run function. Thus, users need to make more modifications to their original version compared with NNI, but it is not necessary to prepare individual files for configuration or search space.
\item[-]Results can be visualized using TensorBoard for both NNI and Tune. Nevertheless, NNI outperforms Tune in the design of Web UI, which is easier to use.
\end{itemize}

\begin{table}
    \centering
    \begin{tabular}{cll}
     \hline
     & \textbf{NNI} & \textbf{Tune}\\
     \hline
     {Application} & \tabincell{l}{Toolkit for AutoML , HPO \\and model compression} & {Only for HPO}\\ 
     \hline
     Deployment & {Local, remote and Dockers} & {Local and remote}\\
     \hline
     {Algorithm} & \tabincell{l}{Support almost all SOTA algorithms\\ except PBT} & \tabincell{l}{Support almost all SOTA algorithms \\including PBT }\\
     \hline
     {Implementation} & \tabincell{l}{By rewriting the algorithm \\in format of NNI} & \tabincell{l}{By importing existing libraries\\ with the interface if Tune} \\
     \hline
     \tabincell{c}{Start \\an Experiment} & \tabincell{l}{Import NNI+ \emph{config. file} \\+ search space file} & \tabincell{l}{Define a trainable class \\+ \emph{tune.run} function} \\
     \hline
     \tabincell{l}{Modification on \\original code} & {Less. By adding individual files} & {More. By build a trainable class}\\
     \hline
     Visualization & \tabincell{l}{Web UI and TensorBoard \\(TensorBoardX)} & \tabincell{l}{TensorBoard \\(TensorBoardX)}\\
     \hline
    \end{tabular}
    \caption{Comparison between closed-source services and open-source toolkits}
    \label{Table 4}
\end{table}

\indent
Tune and NNI also have advantages in common, which are as follows:\\
\begin{itemize}
\item[-]Ease of deployment: Compared to closed-source services, users need to deploy the framework by their own, but both Tune and NNI can be deployed with only one line of command.
\item[-]Scalability: Both of them perform parallel tuning on multi computation units or servers. Users just need to add a few lines in their original code or in a certain file.
\item[-]Support: Support for more state-of-the-art hyper-parameter searching and early stopping algorithms compared with Google Vizier and SageMaker.
\item[-]Tune implements most search algorithms by importing existing libraries (e.g., BayesOpt and HyperOpt), whereas the developers of NNI rewrite most of them. Therefore, users may need to install the relevant package when using a certain algorithm.
\item[-]Open source implies flexibility: Both provide extensible interfaces for algorithms, allowing users to implement new ones. Furthermore, open source implies the lack of the need to pay for all of these features.
\end{itemize}

\indent
However, Tune and NNI also share some drawbacks. They are indeed easy to use as a comprehensive HPO framework, but not as convenient as closed-source toolkits supported by web services. Users cannot avoid deployment and configuration work, and they cannot expect to run thousands of trials simultaneously because of the limitation of resources. Moreover, neither has an interactive GUI; users must use command lines to start adjusting an HPO experiment.\\

\indent
Nevertheless, Tune and NNI are still the most convenient toolkits for hyper-parameter training. In general, they are more suitable for developers with some experience of model training.\\

\section{Discussion and Extensions}
This section concludes the paper with a discussion on the comparison between algorithms (Section 5.1) and evaluations of HPO (Section 5.2).\\
\indent
The concept of HPO has a history of over 20 years~\citep{kohavi95}, and it is a widely used mature technique. In the last few years, with the prosperous development of deep learning, automatic HPO has also become a potential tool to adjust a model for certain applications. Given the scale of deep learning models and their set of hyper-parameters, the key challenges concerning automatic HPO are as follows:\\
\begin{itemize}
\item[-]Many classical search algorithms (e.g., grid search, random search, and Bayesian methods) have become infeasible because of the large number of hyper-parameters and complex configuration space.
\item[-]Evaluation of models with a certain structure or certain choice of hyper-parameters could be extremely expensive.
\item[-]Parallelism will dramatically reduce the total training time. It is a must for HPO tasks, especially for deep models.
\item[-]All in all, computational efficiency is extremely crucial.
\end{itemize}

\subsection{Comparability between different algorithms}
This section makes a brief comparison between major algorithms, including their advantages for saving computational resources, disadvantages in accuracy and efficiency, and their applicability (Table \ref{Table 5}).\\
- \emph{Efforts made by algorithms to save computational resources}\\
\indent
Random search: For deep neural networks, the most important feature of random search is computational parallelism. In addition, it is easy to combine with early stopping strategies, and such combination will largely improve the efficiency for narrowing down the search space.\\
\indent
BO: This is the classical method for optimization addressing the experience of prior search. It is a guided search method with surrogate model and acquisition functions, balancing the exploration and exploitation processes. Bayesian optimization and its variants are the most reliable and convincing methods for optimization. Many recent studies have been conducted to fit this method into the HPO task for deep neural networks.\\
\indent
Multi-armed bandit algorithms: These are actually a series of algorithms based on the non-stochastic infinite-armed bandit problem, including SHA, HB, BOHB, and ASHA. They are different from the abovementioned search algorithms because they introduce an early stopping method to balance the performance and training time. The core schedule for trials is the same for these algorithms: first assign all trials a given initial budget, remove the worse half or a certain percentage, adjust the budget for the remaining trials according to the drop percentage, and repeat until only one trial is left. Trials with poor performance will be terminated halfway instead of trained to convergence to save computation time. These algorithms are more computationally efficient for dealing with increasingly complex models and increasing data size for deep learning networks. Furthermore, they are easy to deploy in parallel.\\
\indent
PBT: This is unique in that it combines hyper-parameter tuning and model trainings. It is a hybrid of random search (parallel optimization) and hand tuning (sequential optimization). PBT starts with several networks trained in parallel, drops trials with the worst performance, and adapts the model by refining hyper-parameters with a certain ratio and inheriting them from the model with good performance. With PBT, a model is trained to have nonstationary hyper-parameters at different stages, with warm start instead of waiting to convergence. Among all the algorithms and schedules, PBT is the most efficient because the parameters and hyper-parameters are trained simultaneously.\\
\begin{table}
    \centering
    \begin{tabular}{clll}
     \hline
     & \textbf{Advantage} & \textbf{Disadvantage} & \textbf{\tabincell{c}{Applicability \\for DNN}} \\
     \hline
     \tabincell{c}{Grid\\ Search} & \tabincell{l}{- Simple\\- Parallelism} & {- “curse of dimensionality”} & \tabincell{l}{- Applicable if only a \\few HPs to tune}\\ 
     \hline
     \tabincell{c}{Random\\ search} & \tabincell{l}{- Parallelism\\- Easy to combine with\\ early stopping methods} & \tabincell{l}{- Low efficiency\\- Cannot promise an\\ optimum} & \tabincell{l}{- Convenient for early \\stage}\\
     \hline
     \tabincell{c}{Bayesian\\optimization} & \tabincell{l}{- Reliable and promising\\- Foundation of many\\ other algorithms} & \tabincell{l}{- Difficult for parallelism \\- Conceptually complex} & \tabincell{l}{- Default algorithm for\\ tools\\- Variants of BO could be\\more applicable (TPE)}\\
     \hline
     \tabincell{c}{Multi-bandit\\methods} & \tabincell{l}{- Conceptually simple \\- Computationally efficient} & \tabincell{l}{- Balance between budget \\ and number of trials} & \tabincell{l}{- Could be a default \\choice\\- Implemented by open-\\sourced libraries.}\\
     \hline
     \tabincell{c}{PBT\\methods} & \tabincell{l}{- Combine HPO and model\\training \\- Parallelism} & \tabincell{l}{- Constant changes to\\computation graph\\- Not extendable to\\ advanced evolution}& \tabincell{l}{- For computationally\\expensive models}\\
     \hline
    \end{tabular}
    \caption{Comparison between major HPO algorithms}
    \label{Table 5}
\end{table}

- \emph{In the meantime, every method has its shortcomings when applied in deep neural networks.}\\
\indent
Grid search: The number of function evaluations grows exponentially with dimensionality and search space. Grid search will quickly become infeasible if the hyper-parameters cannot be limited in a very narrow range.\\
\indent
Random search: Low efficiency occurs, especially when the performance is close to the optimum. Random search only guarantees better results with more resources; it cannot promise to obtain the optimal results, which is consistent with the Monte Carlo method.\\
\indent
BO: The most fatal drawback of the Bayesian method is its deployment in a distributed system. As mentioned previously, parallelism is especially necessary for deep learning, while the feature of borrowing ideas from precious results prevent Bayesian method from direct parallelization. Many recent developments have solved this problem, but it is still not as natural as random search. The conceptual complexity makes the application or improvement of this algorithm challenging.\\
\indent
Multi-armed bandit algorithms: These have their native shortcomings in sampling, but such problems are gradually overcome with the evolution of algorithms. SHA must deal with the tradeoff between budget and number of trials, and drops half every time; HB fixes the first problem by testing several combinations of budgets and trial numbers, and users can determine the scale to drop; and BOHB replaces random search in HB with BO, thereby ensuring both strong anytime performance with early stopping and strong final performance with guided search algorithms. ASHA improves the distributed implementations, and trials can be deployed asynchronously as soon as machines free up. Another unsolved problem is the design of budgets, which is a tricky task in need of user experience of network training and model tuning.\\
\indent
The limitation of PBT lies in the constant changes made to the computation graph. If the model is trained with a pre-defined graph, this process could be complicated. Users can never obtain the “best hyper-parameter set” or a constant value for any hyper-parameter, because neither of them will be stationary during training. The evolutional algorithm is comparatively simple for the original version of PBT; however, it is not quite extendable to advanced evolutional decisions.\\

- \emph{Applicable for deep learning networks?}\\
\indent
Grid search: Theoretically, grid search is not a good method unless only a few hyper-parameters within a narrow spectrum are searched. However, it is still quite popular in practical terms because users tend to narrow down the search space with their knowledge of some hyper-parameters, such as LR. In this case, only several potential values must be tested. In a recent study, the structural parameters determining the width and depth of EfficientNet were achieved using the grid search method ~\citep{tan19b}, which indicated the applicability of this straightforward method.\\
\indent
Random search: Random search is convenient to use for early stages of HPO, especially when combined with an early stopping strategy. When users have no experience in hyper-parameters, they need to narrow down the search space at the first step. The combination of random search and early stopping is efficient in this case because it will explore the whole search space with similar cost to more complicated algorithms. When the search space is narrowed down, random search is not the best choice for a nonguided method.\\
\indent
BO: Many improved versions have been made based on the vanilla BO method for deep neural networks, mainly in the parallelism and configuration spaces. One of the most popular variants is TPE, which is suitable for conditional hyper-parameters and natural for parallelization. It is applied in hyper-parameter tuning for convolutional neural networks and neural language modeling. BO and its variants are implemented in almost every auto-tuning framework, and it is the default search algorithm for Google Vizier and Sagemaker. This indicates that BO is still an active research area for HPO.\\
\indent
Multi-armed bandit algorithms: These could be a default choice for deep neural networks because of their efficiency in computational resources. For a massive parallel system, ASHA outperforms most SOTA algorithms for both language processing and model classification models. In sequential systems and distributed systems with decades of computing units, BOHB could be the first choice for tuning most neural networks and reinforcement learning algorithms because of its efficiency and accuracy. Multi-armed bandit algorithms are implemented in both NNI and Tune.\\
\indent
PBT: This is suitable for computationally expensive models (e.g., generative adversarial networks and transformers) and nonstationary hyper-parameters (e.g., LR). With traditional search algorithms, a model is trained to convergence with different hyper-parameter sets, and then the set with the best performance is chosen. This process could be extremely expensive for large models or large datasets, and PBT is efficient in this case because the hyper-parameters are tuned during training. For nonstationary hyper-parameters, PBT makes a schedule rather than a certain value for them, which is more practical than hand tuning or a man-made schedule.\\

\subsection{Evaluation of results}
\begin{figure}
    \centering
    \includegraphics[width=7.58cm,height=6.79cm]{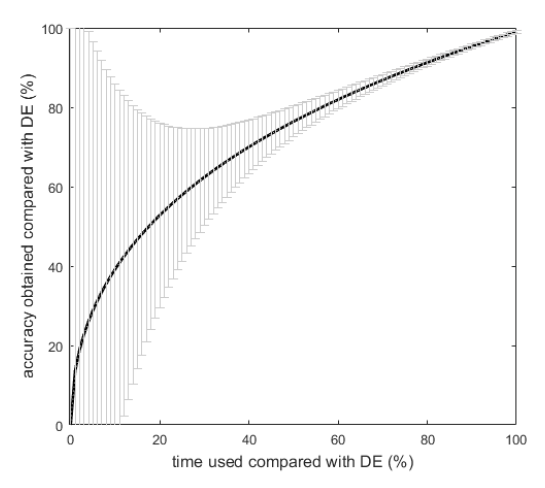}
    \caption{Example of Web UI for NNI}
    \label{Figure 20}
\end{figure}

The previous section mainly discussed the search algorithms for hyper-parameter sets, as well as their feasibility and efficiency. For a large model, the evaluation method could be more critical for saving computation time for unnecessary training because training is usually much more time-consuming compared with the sampling process. An ideal evaluator is supposed be fast, accurate, and simple. However, the choice of evaluation is actually to seek a balance between accuracy and speed (Figure \ref{Figure 20}). In case of deep neural networks, the less time spent on training, the lower accuracy and higher variance there will be.\\
\indent
In the following paragraphs, we review several widely used methods and relevant algorithms for coping with the acceleration of evaluation. The most straightforward and widely used method is a direct evaluation on the target dataset. The direct method is used to evaluate hyper-parameter sets by training a corresponding model to convergence and comparing the loss or accuracy between models. This method is inefficient as it usually takes a long time between two samplings, but it is without doubt the most accurate method. Most users with limited resources cannot afford its expensive time costs, and thus faster methods have been proposed. The most straightforward idea is to decrease the fidelity to trade evaluation time~\citep{march12,fernandez16}.\\
\indent
Furthermore, subsampling has been adapted to deal with time consumption caused by a large training dataset. With this method, the model is trained with a subset of original training data. However, different datasets tend to have different hyper-parameter choices for optimum performance~\citep{kohavi95}; therefore, subsampling is risky in terms of the potential to introduce more noise and uncertainty. A smaller dataset will accelerate the whole process, and the size of the subset must be carefully determined to ensure accuracy. The direct application of a small dataset would definitely generate bias and low fidelity. To solve the problem posed by a small data subset, a transfer series expansion method was proposed on hyper-parameter tuning tasks to accelerate the evaluation with a combination of a set of predictors~\citep{hu19}.\\
\indent
In addition, an early stopping strategy saves training time, as was discussed in Section 6. With an early stopping strategy, the whole training process is divided into sections. At each checkpoint, the model with a performance worse than a certain standard will be terminated. As previously mentioned, median stopping and curve-fitting are typical early stopping methods that can be applied in combination with random search and Bayesian methods among other. Multi-bandit-based algorithms and population-based algorithms are essentially early stopping with different methods for determining checkpoints and judgement criteria. They save training time by terminating poor models halfway, while it is still possible that a model’s performance is worse during the early stage but improves more than remaining ones with more training do. In this case, the model is “killed” too early to be good. Early stopping is the most common method used to save resources, and it is implemented in most AutoML frameworks.\\
\indent
Transfer learning uses knowledge from previous learning tasks to improve the learning or save resources on a target task~\citep{pan09}. The target model is supposed to have a similar structure to the source model~\citep{vanschoren18,yosinski14}; thus, transfer learning is especially feasible for hyper-parameter tuning. Parameter reuse is the most straightforward application of transfer learning in hyper-parameter tuning. The model is initialized with parameters inherited from similar configurations, and is then fine-tuned for further improvement in performance~\citep{yosinski14}. Parameters inherited from similar evaluated configurations could be a feasible initial value to save training time~\citep{bengio07}. The noise introduced by transfer learning is not as predictable as that introduced by subsampling and early stopping. The model could be trapped in different local optima from different initializations, whereas a well-designed initialization performs better with a similar training process~\citep{sutskever13}. Transfer learning has been approved to be workable for tasks with large image datasets, such as ImageNet~\citep{sharif14}. Furthermore, transfer learning is supported by Google Vizier to accelerate the study of warm starting the model training with useful information from relevant studies~\citep{yogatama14}. For a service based on Google Cloud, it is possible to borrow results from other users’ experiments; however, most open-source AutoML frameworks have not implemented a transfer learning module.\\
\indent
Meta learning is also applicable for directly predicting performance of a given model with certain hyper-parameter configurations, by building a surrogate model with previous configurations~\citep{eggensperger15,baker17}. Surrogate models are approximations for making a functional relationship between available data and different types of fidelities~\citep{fernandez16}.  n early works, linear regression was the most common surrogate model~\citep{bensusan01}. With the development of computational resources, more expensive models have also been developed. The curve-fitting method~\citep{domhan15} discussed in subsection 5.2.2 predicted the accuracy with a weighted linear combination. In addition to the most straightforward linear regression, other machine learning algorithms have also been applied as a surrogate model. Guerra made the prediction using support vector machines~\citep{guerra08}; Klein used Bayesian neural networks to build the surrogate model~\citep{klein16}; and Davis the MultiLayer Perceptron to predict the performance of the model with a specific configuration~\citep{davis2018annotative}. This method is especially suitable for HPO among all AutoML tasks because it is easy to use for quantification~\citep{yao18}. The choice of surrogate model may introduce bias because it determines the accuracy for prediction. The accuracy of a surrogate model is influenced by sampling points available for model construction and the accuracy of sampling data~\citep{simpson01}.\\

\section{Conclusion}
This study is motivated by the increasing application of deep neural networks. In this paper, we provided a systematic review on HPO approaches, especially on neural networks. The research began with the introduction of important hyper-parameters for training and structure, and then extended to state-of-the-art search algorithms and schedulers for automated HPO. In addition to outlining related algorithms, we covered services and open-source frameworks, including their pros and cons and applications. Comparisons of algorithms and evaluation methods were further discussed in terms of their feasibility for large models and their consumption of computational resources. This study is dedicated to summarizing information on HPO as a reference for researchers and industrial users.\\


\vskip 0.2in
\bibliography{HPOref}

\begin{thebibliography}{174}
\providecommand{\natexlab}[1]{#1}
\providecommand{\url}[1]{\texttt{#1}}
\expandafter\ifx\csname urlstyle\endcsname\relax
  \providecommand{\doi}[1]{doi: #1}\else
  \providecommand{\doi}{doi: \begingroup \urlstyle{rm}\Url}\fi

\bibitem[hyp()]{hyper}
Hyperparameter: optimization methods and real world model management.
\newblock \emph{URL
  https://missinglink.ai/guides/neural-network-concepts/hyperparameters-optimization-methods-and-real-world-model-management/}.

\bibitem[adv(2018)]{advisor18}
Advisor.
\newblock \url{https:/https://github.com/tobegit3hub/advisor}, 2018.

\bibitem[Abadi et~al.(2016)Abadi, Agarwal, Barham, Brevdo, Chen, Citro,
  Corrado, Davis, Dean, Devin, et~al.]{abadi16}
Mart{\'\i}n Abadi, Ashish Agarwal, Paul Barham, Eugene Brevdo, Zhifeng Chen,
  Craig Citro, Greg~S Corrado, Andy Davis, Jeffrey Dean, Matthieu Devin, et~al.
\newblock Tensorflow: Large-scale machine learning on heterogeneous distributed
  systems.
\newblock \emph{arXiv preprint arXiv:1603.04467}, 2016.

\bibitem[Abbeel(2016)]{abbeel16}
Pieter Abbeel.
\newblock Deep reinforcement learning for robotics.
\newblock
  \url{https://vmayoral.github.io/robots,/ai,/deep/learning,/rl,/reinforcement/learning/2016/07/06/rl-intro/},
  2016.

\bibitem[Agarap(2018)]{agarap18}
Abien~Fred Agarap.
\newblock Deep learning using rectified linear units (relu).
\newblock \emph{arXiv preprint arXiv:1803.08375}, 2018.

\bibitem[Agarwal et~al.(2012)Agarwal, Bartlett, and Duchi]{agarwal12}
Alekh Agarwal, Peter~L Bartlett, and John~C Duchi.
\newblock Oracle inequalities for computationally adaptive model selection.
\newblock \emph{arXiv preprint arXiv:1208.0129}, 2012.

\bibitem[Agrawal(2018)]{agrawal18}
Samarth Agrawal.
\newblock Hyperparameters in deep learning.
\newblock
  \url{https://vmayoral.github.io/robots,/ai,/deep/learning,/rl,/reinforcement/learning/2016/07/06/rl-intro/},
  2018.

\bibitem[Amazon(2018)]{amazon18}
Amazon.
\newblock Advisor.
\newblock \url{https://github.com/tobegit3hub/advisor}, 2018.

\bibitem[Baker et~al.(2017)Baker, Gupta, Raskar, and Naik]{baker17}
Bowen Baker, Otkrist Gupta, Ramesh Raskar, and Nikhil Naik.
\newblock Accelerating neural architecture search using performance prediction.
\newblock \emph{arXiv preprint arXiv:1705.10823}, 2017.

\bibitem[Bello et~al.(2017)Bello, Zoph, Vasudevan, and Le]{bello17}
Irwan Bello, Barret Zoph, Vijay Vasudevan, and Quoc~V Le.
\newblock Neural optimizer search with reinforcement learning.
\newblock In \emph{Proceedings of the 34th International Conference on Machine
  Learning-Volume 70}, pages 459--468. JMLR. org, 2017.

\bibitem[Bengio(2012)]{bengio12}
Yoshua Bengio.
\newblock Practical recommendations for gradient-based training of deep
  architectures.
\newblock In \emph{Neural networks: Tricks of the trade}, pages 437--478.
  Springer, 2012.

\bibitem[Bengio et~al.(2007)Bengio, Lamblin, Popovici, and
  Larochelle]{bengio07}
Yoshua Bengio, Pascal Lamblin, Dan Popovici, and Hugo Larochelle.
\newblock Greedy layer-wise training of deep networks.
\newblock In \emph{Advances in neural information processing systems}, pages
  153--160, 2007.

\bibitem[Bengio et~al.(2011)Bengio, Bastien, Bergeron, Boulanger-Lewandowski,
  Breuel, Chherawala, Cisse, C{\^o}t{\'e}, Erhan, Eustache, et~al.]{bengio11}
Yoshua Bengio, Fr{\'e}d{\'e}ric Bastien, Arnaud Bergeron, Nicolas
  Boulanger-Lewandowski, Thomas Breuel, Youssouf Chherawala, Moustapha Cisse,
  Myriam C{\^o}t{\'e}, Dumitru Erhan, Jeremy Eustache, et~al.
\newblock Deep learners benefit more from out-of-distribution examples.
\newblock In \emph{Proceedings of the Fourteenth International Conference on
  Artificial Intelligence and Statistics}, pages 164--172, 2011.

\bibitem[Bensusan and Kalousis(2001)]{bensusan01}
Hilan Bensusan and Alexandros Kalousis.
\newblock Estimating the predictive accuracy of a classifier.
\newblock In \emph{European Conference on Machine Learning}, pages 25--36.
  Springer, 2001.

\bibitem[Bergstra and Bengio(2012)]{bergstra2012random}
James Bergstra and Yoshua Bengio.
\newblock Random search for hyper-parameter optimization.
\newblock \emph{Journal of machine learning research}, 13\penalty0
  (Feb):\penalty0 281--305, 2012.

\bibitem[Bergstra et~al.(2013)Bergstra, Yamins, and Cox]{bergstra2013making}
James Bergstra, Daniel Yamins, and David~Daniel Cox.
\newblock Making a science of model search: Hyperparameter optimization in
  hundreds of dimensions for vision architectures.
\newblock 2013.

\bibitem[Bergstra et~al.(2011)Bergstra, Bardenet, Bengio, and
  K{\'e}gl]{bergstra11}
James~S Bergstra, R{\'e}mi Bardenet, Yoshua Bengio, and Bal{\'a}zs K{\'e}gl.
\newblock Algorithms for hyper-parameter optimization.
\newblock In \emph{Advances in neural information processing systems}, pages
  2546--2554, 2011.

\bibitem[Bertrand et~al.(2017)Bertrand, Ardon, Perrot, and Bloch]{bertrand17}
Hadrien Bertrand, Roberto Ardon, Matthieu Perrot, and Isabelle Bloch.
\newblock Hyperparameter optimization of deep neural networks: Combining
  hyperband with bayesian model selection.
\newblock In \emph{Conf{\'e}rence sur l’Apprentissage Automatique}, 2017.

\bibitem[Biedenkapp(2018)]{biedenkapp18}
Andr{\'e} Biedenkapp.
\newblock Bohb: robust and efficient hyperparameter optimization at scale.
\newblock
  \url{https://medium.com/criteo-labs/hyper-parameter-optimization-algorithms-2fe447525903},
  2018.

\bibitem[Bissuel(2018)]{bissuel18}
Alo{\"i}s Bissuel.
\newblock Hyper-parameter optimization algorithms: a short review.
\newblock \url{https://www.automl.org/blog_bohb/}, 2018.

\bibitem[Brecque(2018)]{brecque18}
Charles Brecque.
\newblock The intuition behind bayesian optimization with gaussian processes.
\newblock
  \url{https://towardsdatascience.com/the-intuitions-behind-bayesian-optimization-with-gaussian-processes-7e00fcc898a0},
  2018.

\bibitem[Breierova and Choudhari(1996)]{breierova1996introduction}
L~Breierova and M~Choudhari.
\newblock An introduction to sensitivity analysis. massachusetts institute of
  technology.
\newblock Technical report, D-4526-2, 1996.

\bibitem[Brownlee(2016)]{brownlee16}
Jason Brownlee.
\newblock Dropout regularization in deep learning models with keras.
\newblock
  \url{https://machinelearningmastery.com/dropout-regularization-deep-learning-models-keras/},
  2016.

\bibitem[Brownlee(2017)]{brownlee17}
Jason Brownlee.
\newblock Gentle introduction to the adam optimization algorithm for deep
  learning.
\newblock
  \url{https://machinelearningmastery.com/adam-optimization-algorithm-for-deep-learning/},
  2017.

\bibitem[Brownlee(2019)]{brownlee19}
Jason Brownlee.
\newblock How to configure the learning rate when training deep learning neural
  networks.
\newblock
  \url{https://machinelearningmastery.com/learning-rate-for-deep-learning-neural-networks/},
  2019.

\bibitem[Bushaev(2018)]{bushaev18}
Vitaly Bushaev.
\newblock Understanding rmsprop –faster neural network learning.
\newblock
  \url{https://towardsdatascience.com/understanding-rmsprop-faster-neural-network-learning-62e116fcf29a},
  2018.

\bibitem[Chen et~al.(2015)Chen, Li, Li, Lin, Wang, Wang, Xiao, Xu, Zhang, and
  Zhang]{chen15}
Tianqi Chen, Mu~Li, Yutian Li, Min Lin, Naiyan Wang, Minjie Wang, Tianjun Xiao,
  Bing Xu, Chiyuan Zhang, and Zheng Zhang.
\newblock Mxnet: A flexible and efficient machine learning library for
  heterogeneous distributed systems.
\newblock \emph{arXiv preprint arXiv:1512.01274}, 2015.

\bibitem[Chollet(2018)]{chollet18}
Francois Chollet.
\newblock \emph{Deep Learning mit Python und Keras: Das Praxis-Handbuch vom
  Entwickler der Keras-Bibliothek}.
\newblock MITP-Verlags GmbH \& Co. KG, 2018.

\bibitem[Davis and Giraud-Carrier(2018)]{davis2018annotative}
C~Davis and C~Giraud-Carrier.
\newblock Annotative experts for hyperparameter selection.
\newblock In \emph{AutoML Workshop at ICML}, 2018.

\bibitem[Desautels et~al.(2014)Desautels, Krause, and Burdick]{desautels14}
Thomas Desautels, Andreas Krause, and Joel~W Burdick.
\newblock Parallelizing exploration-exploitation tradeoffs in gaussian process
  bandit optimization.
\newblock \emph{Journal of Machine Learning Research}, 15:\penalty0 3873--3923,
  2014.

\bibitem[DeVries and Taylor(2017)]{devries17}
Terrance DeVries and Graham~W Taylor.
\newblock Improved regularization of convolutional neural networks with cutout.
\newblock \emph{arXiv preprint arXiv:1708.04552}, 2017.

\bibitem[Do(2007)]{do07}
Chuong~B Do.
\newblock Gaussian processes.
\newblock \emph{Stanford University, Stanford, CA, accessed Dec}, 5:\penalty0
  2017, 2007.

\bibitem[Domhan et~al.(2015)Domhan, Springenberg, and Hutter]{domhan15}
Tobias Domhan, Jost~Tobias Springenberg, and Frank Hutter.
\newblock Speeding up automatic hyperparameter optimization of deep neural
  networks by extrapolation of learning curves.
\newblock In \emph{Twenty-Fourth International Joint Conference on Artificial
  Intelligence}, 2015.

\bibitem[Duchi et~al.(2011)Duchi, Hazan, and Singer]{duchi11}
John Duchi, Elad Hazan, and Yoram Singer.
\newblock Adaptive subgradient methods for online learning and stochastic
  optimization.
\newblock \emph{Journal of machine learning research}, 12\penalty0
  (Jul):\penalty0 2121--2159, 2011.

\bibitem[Duvenaud(2014)]{duvenaud14}
David Duvenaud.
\newblock \emph{Automatic model construction with Gaussian processes}.
\newblock PhD thesis, University of Cambridge, 2014.

\bibitem[Eberhart and Shi(1998)]{eberhart98}
Russell~C Eberhart and Yuhui Shi.
\newblock Comparison between genetic algorithms and particle swarm
  optimization.
\newblock In \emph{International conference on evolutionary programming}, pages
  611--616. Springer, 1998.

\bibitem[Eggensperger et~al.(2013)Eggensperger, Feurer, Hutter, Bergstra,
  Snoek, Hoos, and Leyton-Brown]{eggensperger13}
Katharina Eggensperger, Matthias Feurer, Frank Hutter, James Bergstra, Jasper
  Snoek, Holger Hoos, and Kevin Leyton-Brown.
\newblock Towards an empirical foundation for assessing bayesian optimization
  of hyperparameters.
\newblock In \emph{NIPS workshop on Bayesian Optimization in Theory and
  Practice}, volume~10, page~3, 2013.

\bibitem[Eggensperger et~al.(2015)Eggensperger, Hutter, Hoos, and
  Leyton-Brown]{eggensperger15}
Katharina Eggensperger, Frank Hutter, Holger Hoos, and Kevin Leyton-Brown.
\newblock Efficient benchmarking of hyperparameter optimizers via surrogates.
\newblock In \emph{Twenty-Ninth AAAI Conference on Artificial Intelligence},
  2015.

\bibitem[Falkner et~al.(2018)Falkner, Klein, and Hutter]{falkner18}
Stefan Falkner, Aaron Klein, and Frank Hutter.
\newblock Bohb: Robust and efficient hyperparameter optimization at scale.
\newblock \emph{arXiv preprint arXiv:1807.01774}, 2018.

\bibitem[Fasshauer and McCourt(2015)]{fasshauer15}
Gregory Fasshauer and Michael McCourt.
\newblock \emph{Kernel-based approximation methods using Matlab}, volume~19.
\newblock World Scientific Publishing Company, 2015.

\bibitem[Fern{\'a}ndez-Godino et~al.(2016)Fern{\'a}ndez-Godino, Park, Kim, and
  Haftka]{fernandez16}
M~Giselle Fern{\'a}ndez-Godino, Chanyoung Park, Nam-Ho Kim, and Raphael~T
  Haftka.
\newblock Review of multi-fidelity models.
\newblock \emph{arXiv preprint arXiv:1609.07196}, 2016.

\bibitem[Feurer and Hutter(2019)]{feurer19}
Matthias Feurer and Frank Hutter.
\newblock Hyperparameter optimization.
\newblock In \emph{Automated Machine Learning}, pages 3--33. Springer, 2019.

\bibitem[Feurer et~al.(2015)Feurer, Klein, Eggensperger, Springenberg, Blum,
  and Hutter]{feurer15}
Matthias Feurer, Aaron Klein, Katharina Eggensperger, Jost Springenberg, Manuel
  Blum, and Frank Hutter.
\newblock Efficient and robust automated machine learning.
\newblock In \emph{Advances in neural information processing systems}, pages
  2962--2970, 2015.

\bibitem[Frazier et~al.(2009)Frazier, Powell, and Dayanik]{frazier09}
Peter Frazier, Warren Powell, and Savas Dayanik.
\newblock The knowledge-gradient policy for correlated normal beliefs.
\newblock \emph{INFORMS journal on Computing}, 21\penalty0 (4):\penalty0
  599--613, 2009.

\bibitem[Frazier(2018)]{frazier18}
Peter~I Frazier.
\newblock A tutorial on bayesian optimization.
\newblock \emph{arXiv preprint arXiv:1807.02811}, 2018.

\bibitem[Garrido-Merch{\'a}n and Hern{\'a}ndez-Lobato(2017)]{garrido17}
Eduardo~C Garrido-Merch{\'a}n and Daniel Hern{\'a}ndez-Lobato.
\newblock Dealing with integer-valued variables in bayesian optimization with
  gaussian processes.
\newblock \emph{arXiv preprint arXiv:1706.03673}, 2017.

\bibitem[Ginsbourger et~al.(2010)Ginsbourger, Le~Riche, and
  Carraro]{ginsbourger10}
David Ginsbourger, Rodolphe Le~Riche, and Laurent Carraro.
\newblock Kriging is well-suited to parallelize optimization.
\newblock In \emph{Computational intelligence in expensive optimization
  problems}, pages 131--162. Springer, 2010.

\bibitem[Ginsbourger et~al.(2011)Ginsbourger, Janusevskis, and
  Le~Riche]{ginsbourger11}
David Ginsbourger, Janis Janusevskis, and Rodolphe Le~Riche.
\newblock Dealing with asynchronicity in parallel gaussian process based global
  optimization.
\newblock 2011.

\bibitem[Girshick(2015)]{girshick15}
Ross Girshick.
\newblock Fast r-cnn.
\newblock In \emph{Proceedings of the IEEE international conference on computer
  vision}, pages 1440--1448, 2015.

\bibitem[Gittins(1974)]{gittins74}
John Gittins.
\newblock A dynamic allocation index for the sequential design of experiments.
\newblock \emph{Progress in statistics}, pages 241--266, 1974.

\bibitem[Golovin et~al.(2017)Golovin, Solnik, Moitra, Kochanski, Karro, and
  Sculley]{golovin17}
Daniel Golovin, Benjamin Solnik, Subhodeep Moitra, Greg Kochanski, John Karro,
  and D~Sculley.
\newblock Google vizier: A service for black-box optimization.
\newblock In \emph{Proceedings of the 23rd ACM SIGKDD international conference
  on knowledge discovery and data mining}, pages 1487--1495, 2017.

\bibitem[Goodfellow et~al.(2014)Goodfellow, Pouget-Abadie, Mirza, Xu,
  Warde-Farley, Ozair, Courville, and Bengio]{goodfellow14}
Ian Goodfellow, Jean Pouget-Abadie, Mehdi Mirza, Bing Xu, David Warde-Farley,
  Sherjil Ozair, Aaron Courville, and Yoshua Bengio.
\newblock Generative adversarial nets.
\newblock In \emph{Advances in neural information processing systems}, pages
  2672--2680, 2014.

\bibitem[Goodfellow et~al.(2016)Goodfellow, Bengio, and
  Courville]{goodfellow16}
Ian Goodfellow, Yoshua Bengio, and Aaron Courville.
\newblock \emph{Deep learning}.
\newblock MIT press, 2016.

\bibitem[Goodfellow et~al.(2013)Goodfellow, Warde-Farley, Mirza, Courville, and
  Bengio]{goodfellow13}
Ian~J Goodfellow, David Warde-Farley, Mehdi Mirza, Aaron Courville, and Yoshua
  Bengio.
\newblock Maxout networks.
\newblock \emph{arXiv preprint arXiv:1302.4389}, 2013.

\bibitem[Guerra et~al.(2008)Guerra, Prud{\^e}ncio, and Ludermir]{guerra08}
Silvio~B Guerra, Ricardo~BC Prud{\^e}ncio, and Teresa~B Ludermir.
\newblock Predicting the performance of learning algorithms using support
  vector machines as meta-regressors.
\newblock In \emph{International Conference on Artificial Neural Networks},
  pages 523--532. Springer, 2008.

\bibitem[Hamby(1994)]{hamby94}
DM~Hamby.
\newblock A review of techniques for parameter sensitivity analysis of
  environmental models.
\newblock \emph{Environmental monitoring and assessment}, 32\penalty0
  (2):\penalty0 135--154, 1994.

\bibitem[Hammond(2017)]{hammond17}
Mark Hammond.
\newblock Deep reinforcement learning in the enterprise: Bridging the gap from
  games to industry, 2017.

\bibitem[Hansen(2016)]{hansen16}
Nikolaus Hansen.
\newblock The cma evolution strategy: A tutorial.
\newblock \emph{arXiv preprint arXiv:1604.00772}, 2016.

\bibitem[Harrington(2018)]{harrington18}
Charlie Harrington.
\newblock Practical guide to hyperparameters optimization for deep learning
  models.
\newblock 2018.

\bibitem[Harrison(2010)]{harrison10}
Robert~L Harrison.
\newblock Introduction to monte carlo simulation.
\newblock In \emph{AIP conference proceedings}, volume 1204, pages 17--21.
  American Institute of Physics, 2010.

\bibitem[He et~al.(2016)He, Zhang, Ren, and Sun]{he16}
Kaiming He, Xiangyu Zhang, Shaoqing Ren, and Jian Sun.
\newblock Deep residual learning for image recognition.
\newblock In \emph{Proceedings of the IEEE conference on computer vision and
  pattern recognition}, pages 770--778, 2016.

\bibitem[Heaton()]{heaton17}
Jeff Heaton.
\newblock The number of hidden layers, 2017.
\newblock \emph{URL https://www. heatonresearch. com/2017/06/01/hidden-layers.
  html}.

\bibitem[Hennig and Schuler(2012)]{hennig12}
Philipp Hennig and Christian~J Schuler.
\newblock Entropy search for information-efficient global optimization.
\newblock \emph{Journal of Machine Learning Research}, 13\penalty0
  (Jun):\penalty0 1809--1837, 2012.

\bibitem[Hern{\'a}ndez-Lobato et~al.(2014)Hern{\'a}ndez-Lobato, Hoffman, and
  Ghahramani]{hernandez14}
Jos{\'e}~Miguel Hern{\'a}ndez-Lobato, Matthew~W Hoffman, and Zoubin Ghahramani.
\newblock Predictive entropy search for efficient global optimization of
  black-box functions.
\newblock In \emph{Advances in neural information processing systems}, pages
  918--926, 2014.

\bibitem[Hinton et~al.(2012{\natexlab{a}})Hinton, Srivastava, and
  Swersky]{hinton12b}
Geoffrey Hinton, Nitsh Srivastava, and Kevin Swersky.
\newblock Neural networks for machine learning.
\newblock \emph{Coursera, video lectures}, 264:\penalty0 1, 2012{\natexlab{a}}.

\bibitem[Hinton et~al.(2006)Hinton, Osindero, and Teh]{hinton06}
Geoffrey~E Hinton, Simon Osindero, and Yee-Whye Teh.
\newblock A fast learning algorithm for deep belief nets.
\newblock \emph{Neural computation}, 18\penalty0 (7):\penalty0 1527--1554,
  2006.

\bibitem[Hinton et~al.(2012{\natexlab{b}})Hinton, Srivastava, Krizhevsky,
  Sutskever, and Salakhutdinov]{hinton12a}
Geoffrey~E Hinton, Nitish Srivastava, Alex Krizhevsky, Ilya Sutskever, and
  Ruslan~R Salakhutdinov.
\newblock Improving neural networks by preventing co-adaptation of feature
  detectors.
\newblock \emph{arXiv preprint arXiv:1207.0580}, 2012{\natexlab{b}}.

\bibitem[Hochreiter and Schmidhuber(1997)]{hochreiter97}
Sepp Hochreiter and J{\"u}rgen Schmidhuber.
\newblock Long short-term memory.
\newblock \emph{Neural computation}, 9\penalty0 (8):\penalty0 1735--1780, 1997.

\bibitem[Hoffman et~al.(2011)Hoffman, Brochu, and de~Freitas]{hoffman11}
Matthew~D Hoffman, Eric Brochu, and Nando de~Freitas.
\newblock Portfolio allocation for bayesian optimization.
\newblock In \emph{UAI}, pages 327--336. Citeseer, 2011.

\bibitem[Hu et~al.(2019)Hu, Yu, Tu, Yang, Chen, and Dai]{hu19}
Yi-Qi Hu, Yang Yu, Wei-Wei Tu, Qiang Yang, Yuqiang Chen, and Wenyuan Dai.
\newblock Multi-fidelity automatic hyper-parameter tuning via transfer series
  expansion.
\newblock In \emph{Proceedings of the AAAI Conference on Artificial
  Intelligence}, volume~33, pages 3846--3853, 2019.

\bibitem[Huang et~al.(2019)Huang, Cheng, Bapna, Firat, Chen, Chen, Lee, Ngiam,
  Le, Wu, et~al.]{huang19}
Yanping Huang, Youlong Cheng, Ankur Bapna, Orhan Firat, Dehao Chen, Mia Chen,
  HyoukJoong Lee, Jiquan Ngiam, Quoc~V Le, Yonghui Wu, et~al.
\newblock Gpipe: Efficient training of giant neural networks using pipeline
  parallelism.
\newblock In \emph{Advances in Neural Information Processing Systems}, pages
  103--112, 2019.

\bibitem[Hutter(2009)]{hutter09}
Frank Hutter.
\newblock \emph{Automated configuration of algorithms for solving hard
  computational problems}.
\newblock PhD thesis, University of British Columbia, 2009.

\bibitem[Hutter et~al.(2011)Hutter, Hoos, and Leyton-Brown]{hutter11}
Frank Hutter, Holger~H Hoos, and Kevin Leyton-Brown.
\newblock Sequential model-based optimization for general algorithm
  configuration.
\newblock In \emph{International conference on learning and intelligent
  optimization}, pages 507--523. Springer, 2011.

\bibitem[Hutter et~al.(2012)Hutter, Hoos, and Leyton-Brown]{hutter12}
Frank Hutter, Holger~H Hoos, and Kevin Leyton-Brown.
\newblock Parallel algorithm configuration.
\newblock In \emph{International Conference on Learning and Intelligent
  Optimization}, pages 55--70. Springer, 2012.

\bibitem[Igel(2014)]{igel14}
Christian Igel.
\newblock No free lunch theorems: Limitations and perspectives of
  metaheuristics.
\newblock In \emph{Theory and principled methods for the design of
  metaheuristics}, pages 1--23. Springer, 2014.

\bibitem[Igel and H{\"u}sken(2000)]{igel00}
Christian Igel and Michael H{\"u}sken.
\newblock Improving the rprop learning algorithm.
\newblock In \emph{Proceedings of the second international ICSC symposium on
  neural computation (NC 2000)}, volume 2000, pages 115--121. Citeseer, 2000.

\bibitem[Jaderberg et~al.(2017)Jaderberg, Dalibard, Osindero, Czarnecki,
  Donahue, Razavi, Vinyals, Green, Dunning, Simonyan, et~al.]{jaderberg17}
Max Jaderberg, Valentin Dalibard, Simon Osindero, Wojciech~M Czarnecki, Jeff
  Donahue, Ali Razavi, Oriol Vinyals, Tim Green, Iain Dunning, Karen Simonyan,
  et~al.
\newblock Population based training of neural networks.
\newblock \emph{arXiv preprint arXiv:1711.09846}, 2017.

\bibitem[Jamieson and Talwalkar(2016)]{jamieson16}
Kevin Jamieson and Ameet Talwalkar.
\newblock Non-stochastic best arm identification and hyperparameter
  optimization.
\newblock In \emph{Artificial Intelligence and Statistics}, pages 240--248,
  2016.

\bibitem[Jia et~al.(2014)Jia, Shelhamer, Donahue, Karayev, Long, Girshick,
  Guadarrama, and Darrell]{jia14}
Yangqing Jia, Evan Shelhamer, Jeff Donahue, Sergey Karayev, Jonathan Long, Ross
  Girshick, Sergio Guadarrama, and Trevor Darrell.
\newblock Caffe: Convolutional architecture for fast feature embedding.
\newblock In \emph{Proceedings of the 22nd ACM international conference on
  Multimedia}, pages 675--678, 2014.

\bibitem[Jin et~al.(2018)Jin, Song, and Hu]{jin18}
Haifeng Jin, Qingquan Song, and Xia Hu.
\newblock Efficient neural architecture search with network morphism.
\newblock \emph{arXiv preprint arXiv:1806.10282}, 2018.

\bibitem[Jones et~al.(1998)Jones, Schonlau, and Welch]{jones98}
Donald~R Jones, Matthias Schonlau, and William~J Welch.
\newblock Efficient global optimization of expensive black-box functions.
\newblock \emph{Journal of Global optimization}, 13\penalty0 (4):\penalty0
  455--492, 1998.

\bibitem[Joseph(2018)]{joseph18}
Rohan Joseph.
\newblock Grid search for model tuning.
\newblock
  \url{https://towardsdatascience.com/grid-search-for-model-tuning-3319b259367e},
  2018.

\bibitem[Karnin et~al.(2013)Karnin, Koren, and Somekh]{karnin13}
Zohar Karnin, Tomer Koren, and Oren Somekh.
\newblock Almost optimal exploration in multi-armed bandits.
\newblock In \emph{International Conference on Machine Learning}, pages
  1238--1246, 2013.

\bibitem[Karpathy et~al.(2016)]{karpathy16}
Andrej Karpathy et~al.
\newblock Cs231n convolutional neural networks for visual recognition.
\newblock \emph{Neural networks}, 1:\penalty0 1, 2016.

\bibitem[Katz et~al.(2016)Katz, Shin, and Song]{katz16}
Gilad Katz, Eui Chul~Richard Shin, and Dawn Song.
\newblock Explorekit: Automatic feature generation and selection.
\newblock In \emph{2016 IEEE 16th International Conference on Data Mining
  (ICDM)}, pages 979--984. IEEE, 2016.

\bibitem[Kaufmann and Garivier(2017)]{kaufmann17}
Emilie Kaufmann and Aur{\'e}lien Garivier.
\newblock Learning the distribution with largest mean: two bandit frameworks.
\newblock \emph{ESAIM: Proceedings and surveys}, 60:\penalty0 114--131, 2017.

\bibitem[Khandelwal(2018)]{khandelwal18}
R~Khandelwal.
\newblock L1 and l2 regularization.
\newblock \emph{Data Driven Investor,[Online]. Available: https://medium.
  com/datadriveninvestor/l1-l2-regularization-7f1b4fe948f2}, 2018.

\bibitem[King et~al.(1995)King, Feng, and Sutherland]{king95}
Ross~D. King, Cao Feng, and Alistair Sutherland.
\newblock Statlog: comparison of classification algorithms on large real-world
  problems.
\newblock \emph{Applied Artificial Intelligence an International Journal},
  9\penalty0 (3):\penalty0 289--333, 1995.

\bibitem[Kingma and Ba(2014)]{kingma14}
Diederik~P Kingma and Jimmy Ba.
\newblock Adam: A method for stochastic optimization.
\newblock \emph{arXiv preprint arXiv:1412.6980}, 2014.

\bibitem[Kizrak(2019)]{kizrak19}
Ayy{\"u}ce Kizrak.
\newblock Comparison of activation functions for deep neural networks.
\newblock
  \url{https://towardsdatascience.com/comparison-of-activation-functions-for-deep-neural-networks-706ac4284c8a?},
  2019.

\bibitem[Klein et~al.(2016)Klein, Falkner, Springenberg, and Hutter]{klein16}
Aaron Klein, Stefan Falkner, Jost~Tobias Springenberg, and Frank Hutter.
\newblock Learning curve prediction with bayesian neural networks.
\newblock 2016.

\bibitem[Koehrsen(2018)]{Koehrsen18}
Will Koehrsen.
\newblock Comparison of activation functions for deep neural networks.
\newblock \url{https://towardsdatascience.com/bayes-rule-applied-75965e4482ff},
  2018.

\bibitem[Kohavi and John(1995)]{kohavi95}
Ron Kohavi and George~H John.
\newblock Automatic parameter selection by minimizing estimated error.
\newblock In \emph{Machine Learning Proceedings 1995}, pages 304--312.
  Elsevier, 1995.

\bibitem[Komer et~al.(2014)Komer, Bergstra, and Eliasmith]{komer14}
Brent Komer, James Bergstra, and Chris Eliasmith.
\newblock Hyperopt-sklearn: automatic hyperparameter configuration for
  scikit-learn.
\newblock In \emph{ICML workshop on AutoML}, volume~9. Citeseer, 2014.

\bibitem[Krivulin et~al.(2005)Krivulin, Guster, and Hall]{krivulin05}
Nikolai~K Krivulin, Dennis Guster, and Charles Hall.
\newblock Parallel implementation of a random search procedure: an experimental
  study.
\newblock In \emph{5th WSEAS International Conference on Simulation, Modeling
  and Optimization (SMO'05), Corfu Island, Greece, August 17-19, 2005}. WORLD
  SCIENTIFIC PUBL CO PTE LTD, 2005.

\bibitem[Krizhevsky et~al.(2012)Krizhevsky, Sutskever, and
  Hinton]{krizhevsky12}
Alex Krizhevsky, Ilya Sutskever, and Geoffrey~E Hinton.
\newblock Imagenet classification with deep convolutional neural networks.
\newblock In \emph{Advances in neural information processing systems}, pages
  1097--1105, 2012.

\bibitem[Kushner(1964)]{kushner64}
Harold~J Kushner.
\newblock A new method of locating the maximum point of an arbitrary multipeak
  curve in the presence of noise.
\newblock 1964.

\bibitem[Lai and Robbins(1985)]{lai85}
Tze~Leung Lai and Herbert Robbins.
\newblock Asymptotically efficient adaptive allocation rules.
\newblock \emph{Advances in applied mathematics}, 6\penalty0 (1):\penalty0
  4--22, 1985.

\bibitem[Lau(2017)]{lau2017learning}
Suki Lau.
\newblock Learning rate schedules and adaptive learning rate methods for deep
  learning.
\newblock \emph{Towards Data Science}, 2017.

\bibitem[LeCun et~al.(1998)LeCun, Bottou, Bengio, and Haffner]{lecun98}
Yann LeCun, L{\'e}on Bottou, Yoshua Bengio, and Patrick Haffner.
\newblock Gradient-based learning applied to document recognition.
\newblock \emph{Proceedings of the IEEE}, 86\penalty0 (11):\penalty0
  2278--2324, 1998.

\bibitem[Li et~al.(2019)Li, Spyra, Perel, Dalibard, Jaderberg, Gu, Budden,
  Harley, and Gupta]{li19}
Ang Li, Ola Spyra, Sagi Perel, Valentin Dalibard, Max Jaderberg, Chenjie Gu,
  David Budden, Tim Harley, and Pramod Gupta.
\newblock A generalized framework for population based training.
\newblock In \emph{Proceedings of the 25th ACM SIGKDD International Conference
  on Knowledge Discovery \& Data Mining}, pages 1791--1799, 2019.

\bibitem[Li et~al.(2015)Li, Karpathy, and Johnson]{li15}
Fei-Fei Li, Andrej Karpathy, and Justin Johnson.
\newblock Convolutional neural networks for visual recognition, 2015.

\bibitem[Li et~al.(2018)Li, Jamieson, Rostamizadeh, Gonina, Hardt, Recht, and
  Talwalkar]{li18}
Liam Li, Kevin Jamieson, Afshin Rostamizadeh, Ekaterina Gonina, Moritz Hardt,
  Benjamin Recht, and Ameet Talwalkar.
\newblock Massively parallel hyperparameter tuning.
\newblock \emph{arXiv preprint arXiv:1810.05934}, 2018.

\bibitem[Li et~al.(2017)Li, Jamieson, DeSalvo, Rostamizadeh, and
  Talwalkar]{li17}
Lisha Li, Kevin Jamieson, Giulia DeSalvo, Afshin Rostamizadeh, and Ameet
  Talwalkar.
\newblock Hyperband: A novel bandit-based approach to hyperparameter
  optimization.
\newblock \emph{The Journal of Machine Learning Research}, 18\penalty0
  (1):\penalty0 6765--6816, 2017.

\bibitem[Li et~al.(2014)Li, Zhang, Chen, and Smola]{li14}
Mu~Li, Tong Zhang, Yuqiang Chen, and Alexander~J Smola.
\newblock Efficient mini-batch training for stochastic optimization.
\newblock In \emph{Proceedings of the 20th ACM SIGKDD international conference
  on Knowledge discovery and data mining}, pages 661--670, 2014.

\bibitem[Liaw et~al.(2018)Liaw, Liang, Nishihara, Moritz, Gonzalez, and
  Stoica]{liaw18}
Richard Liaw, Eric Liang, Robert Nishihara, Philipp Moritz, Joseph~E Gonzalez,
  and Ion Stoica.
\newblock Tune: A research platform for distributed model selection and
  training.
\newblock \emph{arXiv preprint arXiv:1807.05118}, 2018.

\bibitem[Loizou and Richt{\'a}rik(2017)]{loizou17}
Nicolas Loizou and Peter Richt{\'a}rik.
\newblock Momentum and stochastic momentum for stochastic gradient, newton,
  proximal point and subspace descent methods.
\newblock \emph{arXiv preprint arXiv:1712.09677}, 2017.

\bibitem[Lorenzo et~al.(2017)Lorenzo, Nalepa, Kawulok, Ramos, and
  Pastor]{lorenzo17}
Pablo~Ribalta Lorenzo, Jakub Nalepa, Michal Kawulok, Luciano~Sanchez Ramos, and
  Jos{\'e}~Ranilla Pastor.
\newblock Particle swarm optimization for hyper-parameter selection in deep
  neural networks.
\newblock In \emph{Proceedings of the genetic and evolutionary computation
  conference}, pages 481--488, 2017.

\bibitem[Lu et~al.(2019)Lu, Shin, Su, and Karniadakis]{lu19}
Lu~Lu, Yeonjong Shin, Yanhui Su, and George~Em Karniadakis.
\newblock Dying relu and initialization: Theory and numerical examples.
\newblock \emph{arXiv preprint arXiv:1903.06733}, 2019.

\bibitem[Ma et~al.(2018)Ma, Zhang, Zheng, and Sun]{ma18}
Ningning Ma, Xiangyu Zhang, Hai-Tao Zheng, and Jian Sun.
\newblock Shufflenet v2: Practical guidelines for efficient cnn architecture
  design.
\newblock In \emph{Proceedings of the European Conference on Computer Vision
  (ECCV)}, pages 116--131, 2018.

\bibitem[Maladkar(2018)]{maladkar18}
Kishan Maladkar.
\newblock Why is random search better than grid search more machine learning.
\newblock \url{https://towardsdatascience.com/bayes-rule-applied-75965e4482ff},
  2018.

\bibitem[March and Willcox(2012)]{march12}
Andrew March and Karen Willcox.
\newblock Provably convergent multifidelity optimization algorithm not
  requiring high-fidelity derivatives.
\newblock \emph{AIAA journal}, 50\penalty0 (5):\penalty0 1079--1089, 2012.

\bibitem[Masters and Luschi(2018)]{masters18}
Dominic Masters and Carlo Luschi.
\newblock Revisiting small batch training for deep neural networks.
\newblock \emph{arXiv preprint arXiv:1804.07612}, 2018.

\bibitem[Melis et~al.(2017)Melis, Dyer, and Blunsom]{melis17}
G{\'a}bor Melis, Chris Dyer, and Phil Blunsom.
\newblock On the state of the art of evaluation in neural language models.
\newblock \emph{arXiv preprint arXiv:1707.05589}, 2017.

\bibitem[Microsoft(2018)]{msr18}
Microsoft.
\newblock Neural network intelligence.
\newblock \url{https://github.com/microsoft/nni#nni-released-reminder}, 2018.

\bibitem[Mo{\v{c}}kus(1975)]{mockus75}
Jonas Mo{\v{c}}kus.
\newblock On bayesian methods for seeking the extremum.
\newblock In \emph{Optimization techniques IFIP technical conference}, pages
  400--404. Springer, 1975.

\bibitem[Mockus et~al.(1978)Mockus, Tiesis, and Zilinskas]{mockus78}
Jonas Mockus, Vytautas Tiesis, and Antanas Zilinskas.
\newblock The application of bayesian methods for seeking the extremum.
\newblock \emph{Towards global optimization}, 2\penalty0 (117-129):\penalty0 2,
  1978.

\bibitem[Murray(2016)]{murray16}
I~Murray.
\newblock Gaussian processes and kernels, 2016.

\bibitem[nachiket tanksale(2018)]{tanksale18}
nachiket tanksale.
\newblock Finding good learning rate and the one cycle policy.
\newblock
  \url{https://towardsdatascience.com/finding-good-learning-rate-and-the-one-cycle-policy-7159fe1db5d6},
  2018.

\bibitem[Nair and Hinton(2010)]{nair10}
Vinod Nair and Geoffrey~E Hinton.
\newblock Rectified linear units improve restricted boltzmann machines.
\newblock In \emph{Proceedings of the 27th international conference on machine
  learning (ICML-10)}, pages 807--814, 2010.

\bibitem[Ng(2017)]{ng2017improving}
Andrew Ng.
\newblock Improving deep neural networks: Hyperparameter tuning, regularization
  and optimization.
\newblock \emph{Deeplearning. ai on Coursera}, 2017.

\bibitem[Ng(2004)]{ng04}
Andrew~Y Ng.
\newblock Feature selection, l 1 vs. l 2 regularization, and rotational
  invariance.
\newblock In \emph{Proceedings of the twenty-first international conference on
  Machine learning}, page~78, 2004.

\bibitem[Oppermann(2019)]{oppermann19}
Artem Oppermann.
\newblock Overfitting and underfitting in deep learning.
\newblock
  \url{https://www.deeplearning-academy.com/p/ai-wiki-overfitting-underfitting},
  2019.

\bibitem[Orive et~al.(2014)Orive, Sorrosal, Borges, Mart{\'\i}n, and
  Alonso-Vicario]{orive14}
David Orive, G~Sorrosal, Cruz~Enrique Borges, C~Mart{\'\i}n, and
  A~Alonso-Vicario.
\newblock Evolutionary algorithms for hyperparameter tuning on neural networks
  models.
\newblock In \emph{Proceedings of the 26th european modeling \& simulation
  symposium. Burdeos, France}, pages 402--409, 2014.

\bibitem[Pan and Yang(2009)]{pan09}
Sinno~Jialin Pan and Qiang Yang.
\newblock A survey on transfer learning.
\newblock \emph{IEEE Transactions on knowledge and data engineering},
  22\penalty0 (10):\penalty0 1345--1359, 2009.

\bibitem[Paszke et~al.(2019)Paszke, Gross, Massa, Lerer, Bradbury, Chanan,
  Killeen, Lin, Gimelshein, Antiga, et~al.]{paszke19}
Adam Paszke, Sam Gross, Francisco Massa, Adam Lerer, James Bradbury, Gregory
  Chanan, Trevor Killeen, Zeming Lin, Natalia Gimelshein, Luca Antiga, et~al.
\newblock Pytorch: An imperative style, high-performance deep learning library.
\newblock In \emph{Advances in Neural Information Processing Systems}, pages
  8024--8035, 2019.

\bibitem[Provost et~al.(1999)Provost, Jensen, and Oates]{provost99}
Foster Provost, David Jensen, and Tim Oates.
\newblock Efficient progressive sampling.
\newblock In \emph{Proceedings of the fifth ACM SIGKDD international conference
  on Knowledge discovery and data mining}, pages 23--32, 1999.

\bibitem[Qian(1999)]{qian99}
Ning Qian.
\newblock On the momentum term in gradient descent learning algorithms.
\newblock \emph{Neural networks}, 12\penalty0 (1):\penalty0 145--151, 1999.

\bibitem[Ramachandran et~al.(2017{\natexlab{a}})Ramachandran, Zoph, and
  Le]{ramachandran17a}
Prajit Ramachandran, Barret Zoph, and Quoc~V Le.
\newblock Swish: a self-gated activation function.
\newblock \emph{arXiv preprint arXiv:1710.05941}, 7, 2017{\natexlab{a}}.

\bibitem[Ramachandran et~al.(2017{\natexlab{b}})Ramachandran, Zoph, and
  Le]{ramachandran17b}
Prajit Ramachandran, Barret Zoph, and Quoc~V Le.
\newblock Searching for activation functions.
\newblock \emph{arXiv preprint arXiv:1710.05941}, 2017{\natexlab{b}}.

\bibitem[Rasmussen(2003)]{rasmussen03}
Carl~Edward Rasmussen.
\newblock Gaussian processes in machine learning.
\newblock In \emph{Summer School on Machine Learning}, pages 63--71. Springer,
  2003.

\bibitem[Redmon et~al.(2016)Redmon, Divvala, Girshick, and Farhadi]{redmon16}
Joseph Redmon, Santosh Divvala, Ross Girshick, and Ali Farhadi.
\newblock You only look once: Unified, real-time object detection.
\newblock In \emph{Proceedings of the IEEE conference on computer vision and
  pattern recognition}, pages 779--788, 2016.

\bibitem[Ripley(1993)]{ripley93}
Brian~D Ripley.
\newblock Statistical aspects of neural networks.
\newblock \emph{Networks and chaos—statistical and probabilistic aspects},
  50:\penalty0 40--123, 1993.

\bibitem[Robbins and Monro(1951)]{robbins51}
Herbert Robbins and Sutton Monro.
\newblock A stochastic approximation method.
\newblock \emph{The annals of mathematical statistics}, pages 400--407, 1951.

\bibitem[Rodriguez(2018)]{rodriguez18}
Jesus Rodriguez.
\newblock Understanding hyperparameters optimization in deep learning models:
  Concepts and tools, 2018.

\bibitem[Ruder(2016)]{ruder16}
Sebastian Ruder.
\newblock An overview of gradient descent optimization algorithms.
\newblock \emph{arXiv preprint arXiv:1609.04747}, 2016.

\bibitem[Sandler et~al.(2018)Sandler, Howard, Zhu, Zhmoginov, and
  Chen]{sandler18}
Mark Sandler, Andrew Howard, Menglong Zhu, Andrey Zhmoginov, and Liang-Chieh
  Chen.
\newblock Mobilenetv2: Inverted residuals and linear bottlenecks.
\newblock In \emph{Proceedings of the IEEE conference on computer vision and
  pattern recognition}, pages 4510--4520, 2018.

\bibitem[Shahriari et~al.(2014)Shahriari, Wang, Hoffman, Bouchard-C{\^o}t{\'e},
  and de~Freitas]{shahriari14}
Bobak Shahriari, Ziyu Wang, Matthew~W Hoffman, Alexandre Bouchard-C{\^o}t{\'e},
  and Nando de~Freitas.
\newblock An entropy search portfolio for bayesian optimization.
\newblock \emph{arXiv preprint arXiv:1406.4625}, 2014.

\bibitem[Shahriari et~al.(2015)Shahriari, Swersky, Wang, Adams, and
  De~Freitas]{shahriari15}
Bobak Shahriari, Kevin Swersky, Ziyu Wang, Ryan~P Adams, and Nando De~Freitas.
\newblock Taking the human out of the loop: A review of bayesian optimization.
\newblock \emph{Proceedings of the IEEE}, 104\penalty0 (1):\penalty0 148--175,
  2015.

\bibitem[Sharif~Razavian et~al.(2014)Sharif~Razavian, Azizpour, Sullivan, and
  Carlsson]{sharif14}
Ali Sharif~Razavian, Hossein Azizpour, Josephine Sullivan, and Stefan Carlsson.
\newblock Cnn features off-the-shelf: an astounding baseline for recognition.
\newblock In \emph{Proceedings of the IEEE conference on computer vision and
  pattern recognition workshops}, pages 806--813, 2014.

\bibitem[Shiffman et~al.(2012)Shiffman, Fry, and Marsh]{shiffman12}
Daniel Shiffman, Shannon Fry, and Zannah Marsh.
\newblock \emph{The nature of code}.
\newblock D. Shiffman, 2012.

\bibitem[Silver(2014)]{silver14}
David Silver.
\newblock Lecture 9: Exploration and exploitation.
\newblock \emph{Computer Science Department, University of London}, 2014.

\bibitem[Simon(2013)]{simon13}
Dan Simon.
\newblock \emph{Evolutionary optimization algorithms}.
\newblock John Wiley \& Sons, 2013.

\bibitem[Simpson et~al.(2001)Simpson, Poplinski, Koch, and Allen]{simpson01}
Timothy~W Simpson, JD~Poplinski, Patrick~N Koch, and Janet~K Allen.
\newblock Metamodels for computer-based engineering design: survey and
  recommendations.
\newblock \emph{Engineering with computers}, 17\penalty0 (2):\penalty0
  129--150, 2001.

\bibitem[Slivkins et~al.(2019)]{slivkins19}
Aleksandrs Slivkins et~al.
\newblock Introduction to multi-armed bandits.
\newblock \emph{Foundations and Trends{\textregistered} in Machine Learning},
  12\penalty0 (1-2):\penalty0 1--286, 2019.

\bibitem[Smith(2017)]{smith17}
Leslie~N Smith.
\newblock Cyclical learning rates for training neural networks.
\newblock In \emph{2017 IEEE Winter Conference on Applications of Computer
  Vision (WACV)}, pages 464--472. IEEE, 2017.

\bibitem[Snoek et~al.(2012)Snoek, Larochelle, and Adams]{snoek12}
Jasper Snoek, Hugo Larochelle, and Ryan~P Adams.
\newblock Practical bayesian optimization of machine learning algorithms.
\newblock In \emph{Advances in neural information processing systems}, pages
  2951--2959, 2012.

\bibitem[Snoek et~al.(2015)Snoek, Rippel, Swersky, Kiros, Satish, Sundaram,
  Patwary, Prabhat, and Adams]{snoek15}
Jasper Snoek, Oren Rippel, Kevin Swersky, Ryan Kiros, Nadathur Satish,
  Narayanan Sundaram, Mostofa Patwary, Mr~Prabhat, and Ryan Adams.
\newblock Scalable bayesian optimization using deep neural networks.
\newblock In \emph{International conference on machine learning}, pages
  2171--2180, 2015.

\bibitem[Sparks et~al.(2015)Sparks, Talwalkar, Franklin, Jordan, and
  Kraska]{sparks15}
Evan~R Sparks, Ameet Talwalkar, Michael~J Franklin, Michael~I Jordan, and Tim
  Kraska.
\newblock Tupaq: An efficient planner for large-scale predictive analytic
  queries.
\newblock \emph{arXiv preprint arXiv:1502.00068}, 2015.

\bibitem[Srinivas et~al.(2009)Srinivas, Krause, Kakade, and Seeger]{srinivas09}
Niranjan Srinivas, Andreas Krause, Sham~M Kakade, and Matthias Seeger.
\newblock Gaussian process optimization in the bandit setting: No regret and
  experimental design.
\newblock \emph{arXiv preprint arXiv:0912.3995}, 2009.

\bibitem[Srivastava et~al.(2014)Srivastava, Hinton, Krizhevsky, Sutskever, and
  Salakhutdinov]{srivastava14}
Nitish Srivastava, Geoffrey Hinton, Alex Krizhevsky, Ilya Sutskever, and Ruslan
  Salakhutdinov.
\newblock Dropout: a simple way to prevent neural networks from overfitting.
\newblock \emph{The journal of machine learning research}, 15\penalty0
  (1):\penalty0 1929--1958, 2014.

\bibitem[Strauss(2007)]{strauss07}
Martin~J. Strauss.
\newblock Typical steps for solving optimization problems.
\newblock \emph{Electrical Engineering and Computer Science, University of
  Michigan}, 2007.

\bibitem[Strobl et~al.(2008)Strobl, Boulesteix, Kneib, Augustin, and
  Zeileis]{strobl08}
Carolin Strobl, Anne-Laure Boulesteix, Thomas Kneib, Thomas Augustin, and Achim
  Zeileis.
\newblock Conditional variable importance for random forests.
\newblock \emph{BMC bioinformatics}, 9\penalty0 (1):\penalty0 307, 2008.

\bibitem[Sutskever et~al.(2013)Sutskever, Martens, Dahl, and
  Hinton]{sutskever13}
Ilya Sutskever, James Martens, George Dahl, and Geoffrey Hinton.
\newblock On the importance of initialization and momentum in deep learning.
\newblock In \emph{International conference on machine learning}, pages
  1139--1147, 2013.

\bibitem[Swersky et~al.(2014)Swersky, Snoek, and Adams]{swersky14}
Kevin Swersky, Jasper Snoek, and Ryan~Prescott Adams.
\newblock Freeze-thaw bayesian optimization.
\newblock \emph{arXiv preprint arXiv:1406.3896}, 2014.

\bibitem[Szegedy et~al.(2016)Szegedy, Vanhoucke, Ioffe, Shlens, and
  Wojna]{szegedy16}
Christian Szegedy, Vincent Vanhoucke, Sergey Ioffe, Jon Shlens, and Zbigniew
  Wojna.
\newblock Rethinking the inception architecture for computer vision.
\newblock In \emph{Proceedings of the IEEE conference on computer vision and
  pattern recognition}, pages 2818--2826, 2016.

\bibitem[Tan and Le(2019)]{tan19b}
Mingxing Tan and Quoc~V Le.
\newblock Efficientnet: Rethinking model scaling for convolutional neural
  networks.
\newblock \emph{arXiv preprint arXiv:1905.11946}, 2019.

\bibitem[Tan et~al.(2019)Tan, Chen, Pang, Vasudevan, Sandler, Howard, and
  Le]{tan19a}
Mingxing Tan, Bo~Chen, Ruoming Pang, Vijay Vasudevan, Mark Sandler, Andrew
  Howard, and Quoc~V Le.
\newblock Mnasnet: Platform-aware neural architecture search for mobile.
\newblock In \emph{Proceedings of the IEEE Conference on Computer Vision and
  Pattern Recognition}, pages 2820--2828, 2019.

\bibitem[Vanschoren(2018)]{vanschoren18}
Joaquin Vanschoren.
\newblock Meta-learning: A survey.
\newblock \emph{arXiv preprint arXiv:1810.03548}, 2018.

\bibitem[Vaswani et~al.(2017)Vaswani, Shazeer, Parmar, Uszkoreit, Jones, Gomez,
  Kaiser, and Polosukhin]{vaswani2017attention}
Ashish Vaswani, Noam Shazeer, Niki Parmar, Jakob Uszkoreit, Llion Jones,
  Aidan~N Gomez, {\L}ukasz Kaiser, and Illia Polosukhin.
\newblock Attention is all you need.
\newblock In \emph{Advances in neural information processing systems}, pages
  5998--6008, 2017.

\bibitem[Walia(2017)]{walia17}
Anish~Singh Walia.
\newblock Activation functions and it’s types-which is better?
\newblock \emph{Towards Data Science}, 29, 2017.

\bibitem[Wang(2019)]{wang2019vanishing}
CF~Wang.
\newblock The vanishing gradient problem.
\newblock \emph{Towards Data Science}, 2019.

\bibitem[Wang and Perez(2017)]{wang17}
Jason Wang and Luis Perez.
\newblock The effectiveness of data augmentation in image classification using
  deep learning.
\newblock \emph{Convolutional Neural Networks Vis. Recognit}, page~11, 2017.

\bibitem[Wang et~al.(2018)Wang, Xu, and Wang]{wang18}
Jiazhuo Wang, Jason Xu, and Xuejun Wang.
\newblock Combination of hyperband and bayesian optimization for hyperparameter
  optimization in deep learning.
\newblock \emph{arXiv preprint arXiv:1801.01596}, 2018.

\bibitem[Wolpert and Macready(1997)]{wolpert97}
David~H Wolpert and William~G Macready.
\newblock No free lunch theorems for optimization.
\newblock \emph{IEEE transactions on evolutionary computation}, 1\penalty0
  (1):\penalty0 67--82, 1997.

\bibitem[Yan(2016)]{yan16}
Yan.
\newblock L1 norm regularization and sparsity explained for dummies.
\newblock
  \url{https://medium.com/mlreview/l1-norm-regularization-and-sparsity-explained-for-dummies-5b0e4be3938a},
  2016.

\bibitem[Yao et~al.(2018)Yao, Wang, Chen, Dai, Yi-Qi, Yu-Feng, Wei-Wei, Qiang,
  and Yang]{yao18}
Quanming Yao, Mengshuo Wang, Yuqiang Chen, Wenyuan Dai, Hu~Yi-Qi, Li~Yu-Feng,
  Tu~Wei-Wei, Yang Qiang, and Yu~Yang.
\newblock Taking human out of learning applications: A survey on automated
  machine learning.
\newblock \emph{arXiv preprint arXiv:1810.13306}, 2018.

\bibitem[Yiu(2019)]{yiu19}
Tony Yiu.
\newblock The curse of dimensionality.
\newblock
  \url{https://towardsdatascience.com/the-curse-of-dimensionality-50dc6e49aa1e},
  2019.

\bibitem[Yogatama and Mann(2014)]{yogatama14}
Dani Yogatama and Gideon Mann.
\newblock Efficient transfer learning method for automatic hyperparameter
  tuning.
\newblock In \emph{Artificial intelligence and statistics}, pages 1077--1085,
  2014.

\bibitem[Yosinski et~al.(2014)Yosinski, Clune, Bengio, and Lipson]{yosinski14}
Jason Yosinski, Jeff Clune, Yoshua Bengio, and Hod Lipson.
\newblock How transferable are features in deep neural networks?
\newblock In \emph{Advances in neural information processing systems}, pages
  3320--3328, 2014.

\bibitem[You et~al.(2017)You, Gitman, and Ginsburg]{you17}
Yang You, Igor Gitman, and Boris Ginsburg.
\newblock Large batch training of convolutional networks.
\newblock \emph{arXiv preprint arXiv:1708.03888}, 2017.

\bibitem[Zagoruyko and Komodakis(2016)]{zagoruyko16}
Sergey Zagoruyko and Nikos Komodakis.
\newblock Wide residual networks.
\newblock \emph{arXiv preprint arXiv:1605.07146}, 2016.

\bibitem[Zeiler(2012)]{zeiler12}
Matthew~D Zeiler.
\newblock Adadelta: an adaptive learning rate method.
\newblock \emph{arXiv preprint arXiv:1212.5701}, 2012.

\bibitem[Zoph and Le(2016)]{zoph16}
Barret Zoph and Quoc~V Le.
\newblock Neural architecture search with reinforcement learning.
\newblock \emph{arXiv preprint arXiv:1611.01578}, 2016.

\end{thebibliography}

\end{document}